\documentclass[iicol, table]{sn-jnl}
\usepackage{apacite}
\usepackage{booktabs,xltabular,longtable}
\usepackage{tabu}
\usepackage{multicol}
\usepackage{supertabular}
\usepackage{float}

\jyear{2022}%

\theoremstyle{thmstyleone}%
\theoremstyle{thmstyletwo}%

\theoremstyle{thmstylethree}%

\usepackage{soul,framed}

\usepackage{xcolor}
\usepackage{graphicx}
\usepackage{titlesec}

\begin{document}

\title[Article Title]{Context-Driven Detection of Invertebrate Species in Deep-Sea Video}

\author{\fnm{R. Austin} \sur{McEver}}\email{mcever@ucsb.edu}

\author{\fnm{Bowen} \sur{Zhang}}\email{bowen68@ucsb.edu}

\author{\fnm{Connor} \sur{Levenson}}\email{clevenson@ucsb.edu}

\author{\fnm{A S M} \sur{Iftekhar}}\email{iftekhar@ucsb.edu}

\author{\fnm{B. S.} \sur{Manjunath}}\email{manj@ucsb.edu}

\affil{\orgname{University of California, Santa Barbara}, \orgaddress{\postcode{93106}, \state{California}, \country{United States of America}}}


\abstract{Each year, underwater remotely operated vehicles (ROVs) collect thousands of hours of video of unexplored ocean habitats revealing a plethora of information regarding biodiversity on Earth. However, fully utilizing this information remains a challenge as proper annotations and analysis require trained scientists’ time, which is both limited and costly. To this end, we present a Dataset for Underwater Substrate and Invertebrate Analysis (DUSIA), a benchmark suite and growing large-scale dataset to train, validate, and test methods for temporally localizing four underwater substrates as well as temporally and spatially localizing 59 underwater invertebrate species. DUSIA currently includes over ten hours of footage across 25 videos captured in 1080p at 30 fps by an ROV following pre-planned transects across the ocean floor near the Channel Islands of California. Each video includes annotations indicating the start and end times of substrates across the video in addition to counts of species of interest. Some frames are annotated with precise bounding box locations for invertebrate species of interest, as seen in Figure \ref{fig:1}. To our knowledge, DUSIA is the first dataset of its kind for deep sea exploration, with video from a moving camera, that includes substrate annotations and invertebrate species that are present at significant depths where sunlight does not penetrate. Additionally, we present the novel context-driven object detector (CDD) where we use explicit substrate classification to influence an object detection network to simultaneously predict a substrate and species class influenced by that substrate. We also present a method for improving training on partially annotated bounding box frames. Finally, we offer a baseline method for automating the counting of invertebrate species of interest.}



\maketitle

\begin{figure}
    \centering
    \includegraphics[width=0.65\linewidth]{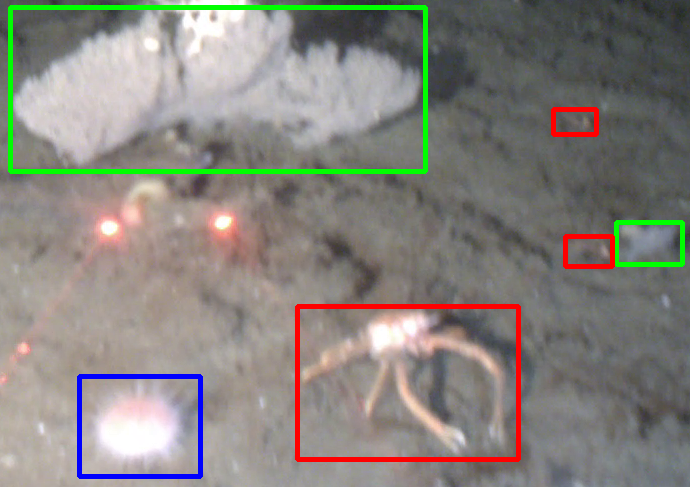}
    \caption{Cropped frame from DUSIA with examples of three classes of interest: fragile pink urchin (blue), gray gorgonian (green), and squat lobster (red). Variations in perspective, occlusion, and size can create large differences across appearances of species individuals and make some individuals, like small squat lobsters, almost invisible, especially in a single frame. Crop shown is 690x487 pixels from a 1920x1080 frame.}
    \label{fig:1}
\end{figure}

\section{Introduction}
\label{sec:intro}

Marine scientists spend enormous amounts of resources on understanding and studying life in our oceans. These studies hold numerous benefits for environmental protection and scientific advancement, including the ability to identify areas of the ocean where certain habitats and substrates exist and where certain species gather. 

A common method for studying underwater habitats consists of planning underwater routes, called transects, then following those paths and recording the environment either by a diver with a camera or using an underwater ROV \cite{shester2017exploring, drap2015rov}. Once the transects have been recorded and videos matched with their GPS locations, common annotation methods require researchers to review each video several times, annotating the substrates that the camera passes over in the first few annotation passes, then counting invertebrates in another pass, and then counting fish species in a final pass to give a better idea of where in the ocean which substrates exist and where different species live. This information is vital to determining species hotspots and finding ways to protect the environment while also meeting human needs for usage of our oceans. These studies ultimately lead to new discoveries as they facilitate exploration of unknown oceanic regions. Currently, however, the sheer amount of data researchers collect can be overwhelmingly expensive and difficult to annotate and utilize as their annotation methods' multiple passes can push annotations times to many times the duration of the video.

Computer vision and machine learning models can significantly aid in managing, utilizing, analyzing, and understanding these videos, ultimately reducing the overall costs of these studies and freeing researchers from tedious annotation tasks. However, developing and training these models require annotated data. Further, the types of annotations generated and used by domain scientists do not directly correspond with the typical types of annotations generated and used by computer vision researchers, requiring new approaches to learning from video data and their annotations.

As a step toward advancement in efficiently computationally analyzing videos from a marine science setting, we introduce DUSIA, a real world scientific dataset including videos collected and annotated by marine scientists who directly use a superset of these videos to advance their own research and exploration. To our knowledge, DUSIA is the first public dataset to contain videos recorded in this challenging moving-camera setting where an underwater ROV drives and records over the ocean floor. This dataset allows us to create solutions to a  host of difficult computer vision problems that have not yet been explored such as classifying and temporally localizing underwater habitats and substrates, counting and tracking invertebrate species as they appear in ROV video, and using these explicit substrate and habitat classifications to help detect and classify invertebrate species. Further, the types of annotations provided in DUSIA differ from those of typical computer vision datasets, requiring new approaches to learning.


%

Our contributions can be summarized as follows:
 \begin{itemize}
  \item DUSIA provides the first publicly available dataset of annotated, full-length videos captured via an underwater ROV. DUSIA's videos are annotated by expert marine scientists with temporal labels indicating substrates, count labels for 59 invertebrate species, partial bounding box labels for ten invertebrate species of interest in the training set, and full bounding box labels for those species of interest in the validation and testing sets.
  \item We introduce the novel Context-Driven Detector (CDD), which uses implicit context representations and explicit context labels to improve bounding box detections. In our case, context refers to explicit class labels of the background. Specifically, our context labels describe the substrate present on the ocean floor, which determine the environment and habitat in which the organisms live. In natural images, context might refer to indoor vs outdoor images or subcategories within such as school, office, library, or supermarket.
  \item We propose Negative Region Dropping, an approach for improving performance of an object detector trained on a dataset with partially annotated images.
  \item Finally, we offer a baseline method for counting invertebrate species individuals in this challenging setting using a detection plus tracking pipeline.
   \end{itemize}

In Section \ref{sec:related} we review other datasets and methods with similar data and highlight how DUSIA differs from previous datasets. Next, in Section \ref{sec:dataset} we discuss the contents and collection of DUSIA's data and annotations. Section \ref{sec:tasks} describes some of the tasks for which DUSIA can be used, and Section \ref{sec:methods} discusses our approaches to those tasks including the novel CDD, Negative Region Dropping, and baseline tracking method. Section \ref{sec:experiments} describes our experiments and results, and Section \ref{sec:discussion} discusses our findings.

\section{Related Works}
\label{sec:related}

Current achievements by deep learning-based vision models do not translate well when it comes to analyzing underwater animals and habitats as there exists a scarcity of well-annotated underwater data. Although there are a few efforts from the computer vision community to collect and annotate underwater data \cite{pedersen2019detection,king2018comparison,boom2014research, marini2018tracking,joly2014lifeclef}, it is hardly enough to tackle this daunting problem, and few of these efforts collect data in the same way or provide annotations for the same goals. In general, collecting underwater image or video data is far more difficult than land data and day to day images of common objects. 
As a result, the whole data collection process becomes complicated and expensive. DUSIA aims to be a collaborative, comprehensive effort to guide the exploration and automated analysis of underwater ecosystems. 
\begin{figure}
    \centering
    \includegraphics[width=0.95\linewidth]{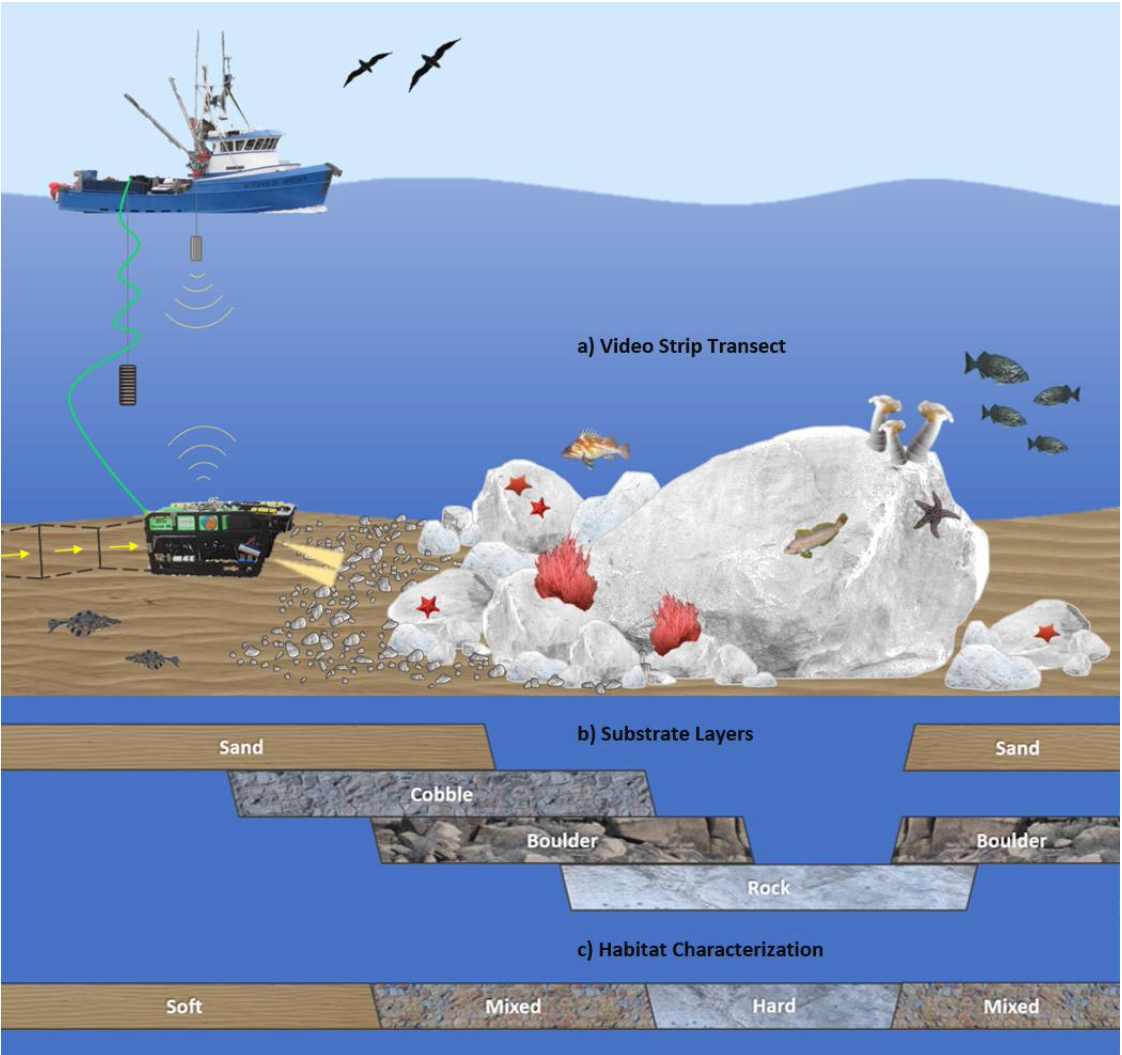}
    \caption{Illustration of the ROV attached to the catamaran, substrate layers, and habitat characterization. Substrates are divided into soft (mud, cobble and sand), hard (rock and boulder), or mixed (a combination of any soft and hard substrates). Illustration courtesy of Marine Applied Research and Exploration (MARE) Group.}
    \label{fig:ROV}
\end{figure} 
\subsection{Underwater Marine Datasets}
Many of the existing underwater marine datasets are developed in order to detect and recognize the various behaviors or simply presence of fish ~\cite{konovalov2019underwater,maaloy2019spatio, boom2014research,joly2014lifeclef,levy2018automated}. Numerous current works ~\cite{konovalov2019underwater,maaloy2019spatio, levy2018automated,ditria2020automating} have validated their fish detection and fish behavior recognition models on these datasets. Interestingly, these methods mainly focus on developing novel data-hungry algorithms, but the data on which the algorithms perform is limited by its static perspective. For example, Maaloy et al  ~\cite{maaloy2019spatio} proposed a dual spatial-temporal recurrent network, but the algorithm is trained and tested on a dataset that is constrained by having no camera movement and working in a covered area. Similarly, Konovalov et al ~\cite{konovalov2019underwater} augments their dataset of underwater fish images with the underwater non-fish images from VOC2012 ~\cite{everingham2015pascal} by restricting their model to generating only binary (fish vs. no fish) predictions. In the same way, ~\cite{ditria2020automating,levy2018automated} confined their models to do analysis only on one single fish. In contrast, DUSIA provides dynamic, high definition ROV video showcasing a rich and varied environment with many species occurring in intermingling groups.

\def \supfih {1.5cm}
\begin{figure*}[t!]
    \centering
    \includegraphics[height=\supfih{}]{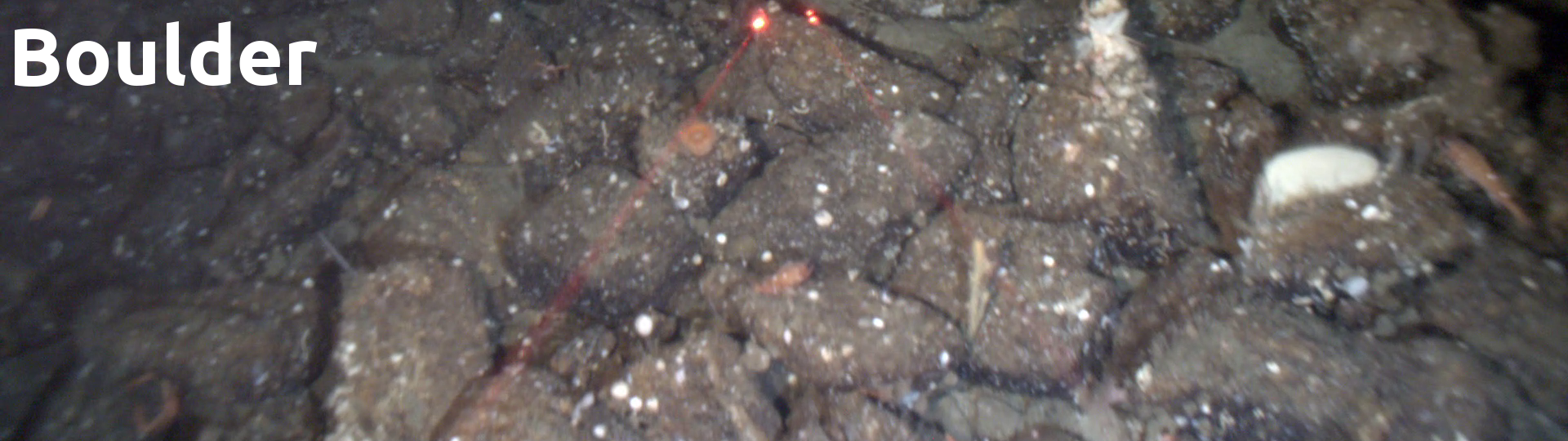}
    \includegraphics[height=\supfih{}]{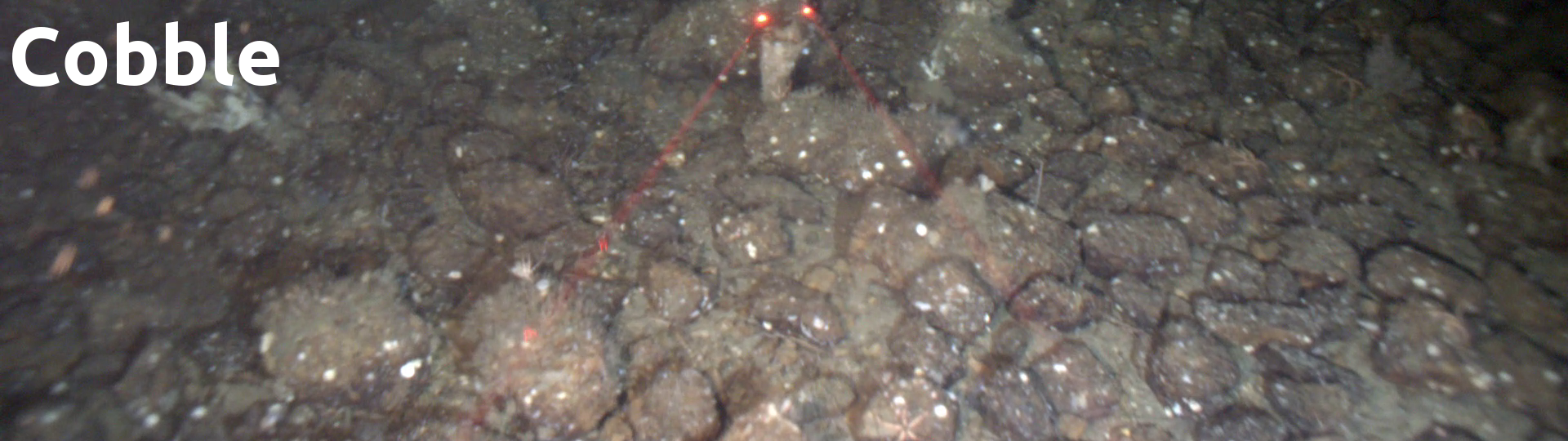}
    \includegraphics[height=\supfih{}]{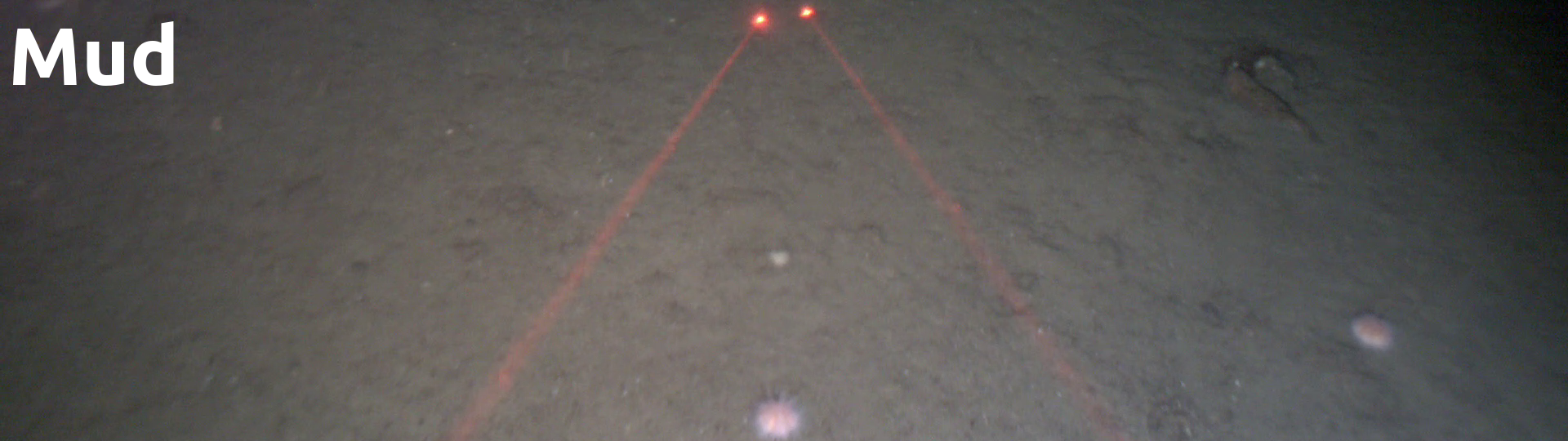}
    \includegraphics[height=\supfih{}]{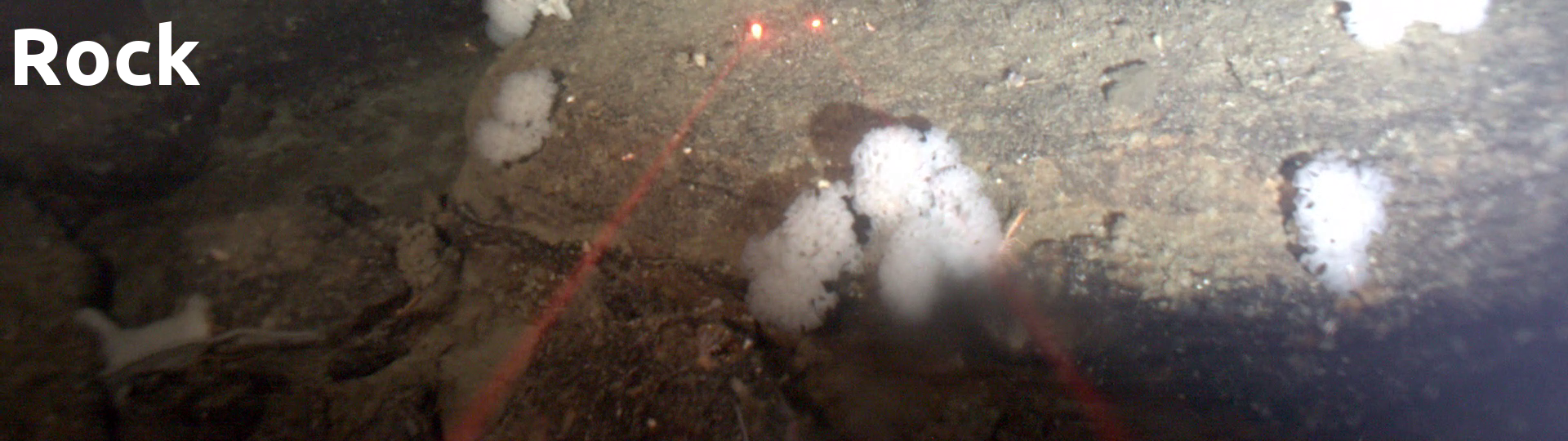}
    \caption{Example frames each containing just one substrate each, indicated by the in frame text}
    \label{fig:subs}
\end{figure*}

Additionally, unlike existing datasets, a novel feature of DUSIA is the utilization of explicit, human-annotated, contextual information such as substrates or habitat in the analysis workflow. Such contextual information can play a vital role in making accurate predictions, especially in the case of identifying fish or other marine animals. Recently, ~\cite{rashid2020trillion} has developed a large scale dataset for habitat mapping using both RGB images and hyperspectral images. This dataset contains a large number of annotated images for classifying different coral reef habitats, but marine animal information is not included in this dataset. DUSIA, in contrast, is unique in this aspect, as it has both explicit substrate and invertebrate annotations.

\subsection{Methodologies}
As mentioned in the previous section, recently, different works have developed deep learning-based algorithms to detect marine species (mostly fishes). Li et al ~\cite{li2015fast} uses a Fast-RCNN ~\cite{girshick2015fast} based network to classify twelve different species of fish. Salman et al ~\cite{salman2016fish} present a deep network to detect fish in 32x32 size video frames. Siddiqui et al ~\cite{siddiqui2018automatic} use a pre-trained object detection CNN network as a generalized feature extractor. The extracted features are then fed to an SVM (support vector machine) for classification of fish. Our baseline method aims to alleviate some of these methods' shortcomings by using explicit substrate predictions to enhance species detections.

\section{Dataset} \label{sec:dataset}
DUSIA consists of over 10 hours of footage captured from preplanned transects along the ocean floor near the Channel Islands of California. This includes 25 HD videos recorded using RGB video cameras attached to an observation class ROV equipped with multiple lighting fixtures recording at depths between 100 and 400 meters. Three of the 25 videos do not contain species of interest, so they are excluded from experiments presented in this paper. DUSIA's videos are part of a large collection, and we plan to release more similar videos from different excursions in the future. 

\subsection{Data Collection}
Surveys of wildlife on the ocean floor generally start with planning a group of paths, called transects, across some region in order to efficiently cover and survey one section of the ocean \cite{shester2017exploring}; however, to protect these fragile ecosystems, DUSIA does not make specific GPS coordinates publicly available. 

Some surveys use scuba divers to collect video along transects, but DUSIA covers larger, deeper areas using an ROV attached to a 77-foot catamaran. During the collection process, the ROV is attached via cable to the catamaran. Once the boat arrives near the beginning of the desired transects, the ROV is placed in the water and remains on a long leash attached to the boat such that the catamaran can follow the transects roughly while the ROV follows its path more precisely via inputs from a remote operator on the boat who makes use of the ROV’s cameras, lights, GPS, and other instruments that indicate the ROV’s location relative to the boat, which allows for computing its GPS location. Figure \ref{fig:ROV} roughly illustrates the ROV rig used for data collection.



\subsection{Substrate Classes and Annotations}
After the collection stage, researchers return to a laboratory where they review, analyze, and annotate each video. DUSIA includes four different substrates: boulder, cobble, mud, and rock. An illustration of each one is shown in Figure \ref{fig:ROV} and frames from the dataset are shown in Figure \ref{fig:subs}. The difference between each depends on the nature of the material makeup of the ocean floor. A description of each substrate can be found in Table \ref{tab:sub_des}, and Table \ref{tab:an_ex} shows a toy example of the annotation format.
  
\begin{table}[t]
\centering
\begin{tabular}{p{0.13\linewidth} p{0.81\linewidth}}
\toprule
Substrate & Description                                                                                                           \\ \midrule
Boulder   & rocky substrate larger than 25 cm in diameter that is detached and clearly movable                                    \\
Cobble    & rocky substrate that is 6 to 25 cm in diameter                                                                        \\
Mud       & very fine sediments that stay suspended in the water when disturbed (loss of visibility)                              \\
Rock      & consolidated rocky substrates that appear attached to the bottom and not movable                                      \\
\bottomrule
\end{tabular}
\caption{Description of the four substrates present in DUSIA}
\label{tab:sub_des}
\end{table}

Each of these substrates may overlap such that a given frame can have multiple substrate labels if enough of multiple substrates are visible. The annotation process includes multiple passes, one for each substrate, where the annotators indicate the start and end times of each substrate occurrence. This arduous process can be alleviated by our methods. 

\subsection{Invertebrate Classes and Annotations} \label{sec:invert_annos}
Once the substrate annotations are completed, scientists make yet another pass over each video, this time annotating invertebrate species, often referencing substrate labels as certain species have a tendency to occur in certain substrates. When a group or individual of a species touches the bottom of the video frame, they pause the video, count the species touching the bottom of the frame, and make note of the time stamp at which the count occurred, giving domain researchers insight into where in the video, in the ocean, and in which substrate, each species tends to occur. We refer to these labels as CABOF, Count At the Bottom of the Frame, labels. 



Count labels provide guidance in learning to classify and detect invertebrate species, they ensure that species individuals are not counted multiple times, and a human could use these labels to learn to label further videos. However, current computer vision methods struggle with weak supervision, and count labels of this nature are unusual for current machine learning methods. 

\begin{table}[ht]
\centering
\begin{tabular}{@{}lrrl@{}}
\toprule
Annotation & \multicolumn{1}{l}{Begin} & \multicolumn{1}{l}{End} & Count                 \\ \midrule
Boulder    & 0:00:20                   & 0:00:25                 &                       \\
FPU        & 0:00:21                   & \multicolumn{1}{l}{}    & \multicolumn{1}{r}{2} \\
Cobble     & 0:00:23                   & 0:01:30                 &                       \\
Mud        & 0:00:40                   & 0:01:20                 &                       \\
SL         & 0:00:49                   & \multicolumn{1}{l}{}    & \multicolumn{1}{r}{1} \\
SL         & 0:00:51                   & \multicolumn{1}{l}{}    & \multicolumn{1}{r}{3} \\
Rock       & 0:01:00                   & 0:03:50                 &                       \\
Mud       & 0:02:10                   & 0:02:15                 &                       \\ \bottomrule
\end{tabular}
\caption{Example of combined substrate and CABOF, Count At the Bottom of the Frame, annotations. Substrates are labeled with beginning and end times, and invertebrate CABOF labels include a single timestamp shown in the Begin column and count.}
\label{tab:an_ex}
\end{table}

\begin{figure*}[t!]
    \centering
    \includegraphics[width=1\linewidth]{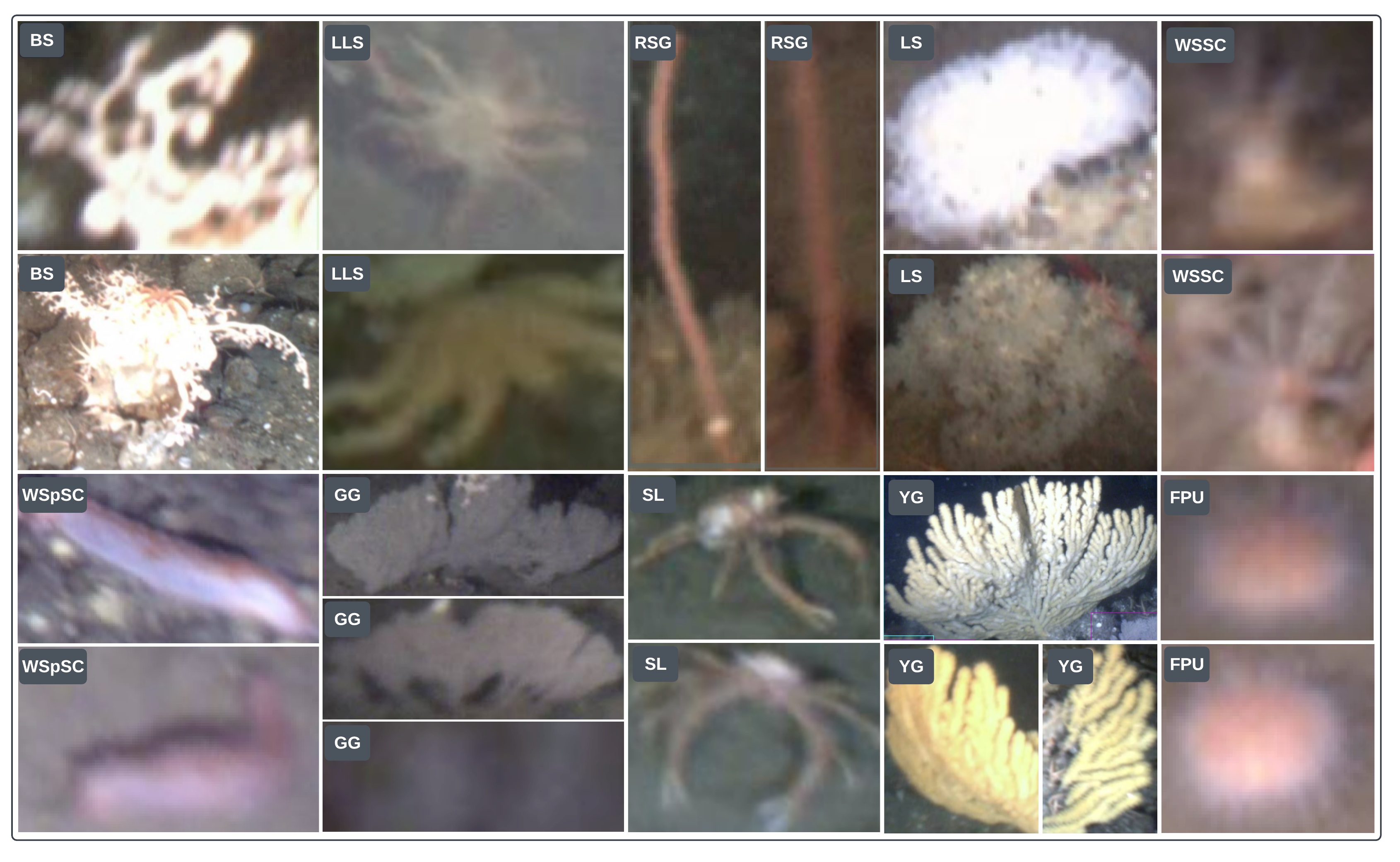}
    \caption{Cropped screenshots of each of the ten species of interest: basket star (BS), fragile pink urchin (FPU), gray gorgonian (GG), long-legged sunflower star (LLS), red swifita gorgonian (RSG), squat lobster (SL), laced sponge (LS), white slipper sea cucumber (WSSC), white spine sea cucumber (WSpSC), and yellow gorgonian (YG).}
    \label{fig:species}
\end{figure*}
\subsubsection{Bounding Box Labels} To address this difficulty, we further annotate a subset of the dataset with bounding box tracks to help enable current computer vision methods, which often require bounding boxes for training and testing, and to validate those methods on DUSIA, using the marine scientists' CABOF labels. First, we select a subset of species to annotate with stronger annotations. We choose ten species, each visualized in Figure \ref{fig:species} because they are some of the most abundant species in the dataset. Appendix \ref{sec:all-spec} shows the counts of all invertebrate species annotated with count labels across DUSIA.

To generate our training set, we randomly select a subset of frames containing count labels for our species of interest. We seek to those frames and back up in the video until the annotated species individual or group, i.e. our annotation target(s), is either in the top half of the screen or first appearing. In the ROV viewpoint, objects typically appear at the top of the frame as the ROV moves forward. Once we back up sufficiently far, we then draw a bounding box or boxes on the annotated target(s), ignoring other instances of species of interest (thus creating partial annotations) due to annotation budget and visibility constraints.

We then jump 10-30 frames at a time adjusting the box location for the annotation target(s) in each frame we land on, referred to as \emph{keyframes}. This process allows for efficient annotation and allows us to interpolate box locations between keyframes for additional annotation points.

The result of this annotation process is a partially annotated training set for learning to detect and later count species of interest. These annotations are partial because we did not attempt to always label every individual of each species of interest in the training set. Instead, we focused only on the annotation targets. Because some individuals of the ten species of interest may be labelled while other individuals of the ten species may not be, we consider these partial labels. 

We chose to partially annotate the our training set so that we could collect boxes tracking each species. In populated areas, there are many species hiding, coming, and going, making collecting full annotations extremely difficult, especially across many frames.

Additionally, we provide some fully annotated frames where we guarantee that all individuals of the ten species of interest in the bottom half of each frame are labelled with a bounding box. We were constrained to the bottom half of the frame due to darkness, murky waters, low visibility, and text embedded in the videos during the collection process. Therefore, we use only the fully annotated bottom half of the validation and testing frames when presenting our detection results. Seeing as the marine scientists count the creatures that touch the bottom of the frame, we expect the bottom half of the frame to provide a good metric for count estimations. These frames are provided for validation and testing.

In order to generate these fully annotated validation and testing frames, we randomly selected a subset of count annotated frames in the validation and test sets. For each of those selected frames, we labelled all instances of species of interest in the bottom half of the frame including but not limited to the original targets. For rare species, we often labelled frames a second or two before and/or after the count annotated frame in order to provide more validation and testing frames. Still, the number of validation and testing frames is limited by the difficulty in collecting these fully annotated frames as well as the scarcity of some species. 

\begin{figure*}[t]
    \centering
    \includegraphics[width=1\linewidth]{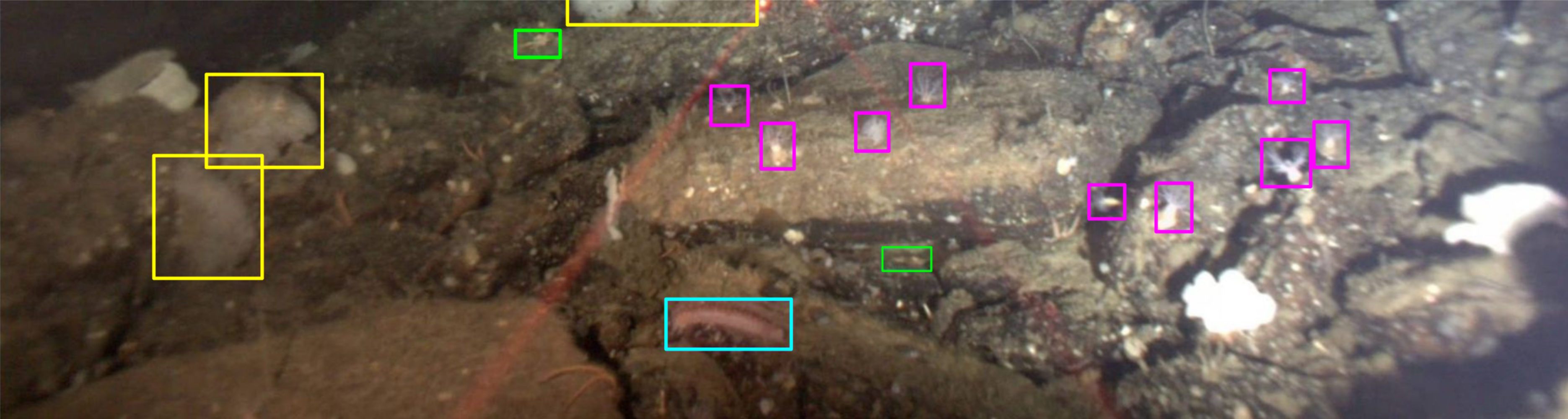}
    \caption{Fully annotated frame example. Color to species map is as follows: yellow: laced sponge, magenta: white spine sea cucumber, cyan: white slipper sea cucumber, green: squat lobster.}
    \label{fig:full}
\end{figure*}

These fully annotated frames took on average 146.5 seconds per frame for trained individuals to annotate. For reference, it took annotators approximately 22.1 seconds per image to fully annotate with single point annotation and 34.9 seconds per image with squiggle supervision in the VOC2012 natural image dataset of 20 classes including cats, busses, and similar common object classes \cite{bearman2016s}. Collecting bounding boxes, consisting of two precise points, with half the number of classes should take a similar amount of time, but the difference in time spent per image illustrates the challenge of annotating DUSIA as each annotator struggled to find every object of interest even after being trained to specifically to localize the species of interest. An example of a fully labelled validation frame is shown in Figure \ref{fig:full}.

\subsection{Dataset Splits}
\label{sec:splits}
We provide a split of the dataset into training, validation, and testing sets with 13, 3, and 6 videos in each split respectively. The training set includes 8,682 keyframes used for training the detector (described in detail in Section \ref{sec:invert_annos}). The validation and test sets respectively include 514 and 677 frames with fully annotated lower halves. Between each split, we attempted to maintain a relatively even distribution across our species of interest; however, preserving this distribution leads to a slightly uneven distribution of substrate occurrences.

\subsection{Statistical Analysis of Data}

Table \ref{tab:sub_ac_sp} shows the frequency of each of the substrate classes present in our dataset. 

Table \ref{tab:spec_ac_sp} shows the frequency of bounding box labels for invertebrate species of interest represented in our dataset, and Table \ref{tab:count_ac_sp} illustrates the frequency of CABOF labels for invertebrate species.

Table \ref{tab:spec_ac_sub} illustrates the distributions of CABOF labels for each species across the different substrates. While not weighted against the relative presence of each substrate, this table still illustrates that certain species occur much more frequently in certain substrates. For example, fragile pink urchins (FPU) rarely occur in the boulder substrate, and frequently occur in mud while laced sponges (LS) almost always occur in a substrate that includes rock. These correlations suggest that learning to predict substrate may aid in learning the relationship between substrate and species and motivate a context driven approach for species detection and counting.

\begin{table}[t!]
\centering
\begin{tabular}{@{}lrrrrr@{}}
\toprule
      & \multicolumn{1}{l}{B} & \multicolumn{1}{l}{C} & \multicolumn{1}{l}{M} & \multicolumn{1}{l}{R} & \multicolumn{1}{l}{Total} \\ \midrule
Train & 70,248                & 247,764               & 259,535               & 183,020               & 760,567                   \\
Val   & 14,899                & 28,694                & 23,656                & 63,322                & 130,571                   \\
Test  & 30,742                & 91,695                & 102,422               & 87,399                & 312,258                   \\
Total & 115,889               & 368,153               & 385,613               & 333,741               & 1,203,396                 \\ \bottomrule
\end{tabular}
\caption{Distribution of number of frames containing each substrate across DUSIA and its splits}
\label{tab:sub_ac_sp}
\end{table}

\begin{table*}[ht]
\centering
\begin{tabular}{lrrrrrrrrrrr}
\hline
      & \multicolumn{1}{l}{BS} & \multicolumn{1}{l}{FPU} & \multicolumn{1}{l}{GG} & \multicolumn{1}{l}{LLS} & \multicolumn{1}{l}{RSG} & \multicolumn{1}{l}{SL} & \multicolumn{1}{l}{LS} & \multicolumn{1}{l}{WSSC} & \multicolumn{1}{l}{WSpSC} & \multicolumn{1}{l}{YG} & \multicolumn{1}{l}{Total} \\ \hline
Train & 1,247                  & 3,675                   & 3,294                  & 735                      & 775                     & 3,264                  & 1,071                  & 1,397                    & 819                       & 1,024                  & 17,301                    \\
Val   & 61                     & 394                     & 259                    & 20                       & 85                      & 594                    & 91                     & 439                      & 51                        & 38                     & 2,032                     \\
Test  & 124                    & 653                     & 277                    & 61                       & 79                      & 1,181                  & 98                     & 506                      & 28                        & 180                    & 3,187                     \\
Total & 1,432                  & 4,722                   & 3,830                  & 816                      & 939                     & 5,039                  & 1,260                  & 2,342                    & 898                       & 1,242                  & 22,520                    \\ \hline
\end{tabular}
\caption{Distribution of bounding box annotations of each species across splits. Note that one species individual may be annotated with multiple bounding boxes as it occurs across multiple frames.}
\label{tab:spec_ac_sp}
\end{table*}

\begin{table*}[ht]
\centering
\begin{tabular}{@{}lrrrrrrrrrrr@{}}
\toprule
      & \multicolumn{1}{l}{BS} & \multicolumn{1}{l}{FPU} & \multicolumn{1}{l}{GG} & \multicolumn{1}{l}{LLS} & \multicolumn{1}{l}{RSG} & \multicolumn{1}{l}{SL} & \multicolumn{1}{l}{LS} & \multicolumn{1}{l}{WSSC} & \multicolumn{1}{l}{WSpSC} & \multicolumn{1}{l}{YG} & \multicolumn{1}{l}{Total} \\ \midrule
Train & 292                    & 2,828                   & 398                    & 269                     & 190                     & 1,649                  & 517                      & 832                      & 279                       & 103                    & 7,357                     \\
Val   & 17                     & 154                     & 80                     & 8                       & 19                      & 208                    & 40                       & 164                      & 22                        & 9                      & 721                       \\
Test  & 52                     & 420                     & 78                     & 29                      & 48                      & 742                    & 75                       & 317                      & 17                        & 38                     & 1,816                     \\
Total & 361                    & 3,402                   & 556                    & 306                     & 257                     & 2,599                  & 632                      & 1,313                    & 318                       & 150                    & 9,894                     \\ \bottomrule
\end{tabular}
\caption{Distribution of CABOF labels across DUSIA and its splits. As described in Section \ref{sec:invert_annos}, each species individual is counted only once when it touches the bottom of the frame.}
\label{tab:count_ac_sp}
\end{table*}

\begin{table*}[ht!]
\centering
\begin{tabular}{lrrrrrrrrrr}
\hline
  & \multicolumn{1}{l}{BS} & \multicolumn{1}{l}{FPU} & \multicolumn{1}{l}{GG} & \multicolumn{1}{l}{LLS} & \multicolumn{1}{l}{RSG} & \multicolumn{1}{l}{SL} & \multicolumn{1}{l}{LS} & \multicolumn{1}{l}{WSSC} & \multicolumn{1}{l}{WSpSC} & \multicolumn{1}{l}{YG} \\ \hline
B & 0.302                  & 0.059                   & 0.362                  & 0.206                    & 0.198                   & 0.219                  & 0.168                  & 0.224                    & 0.176                     & 0.340                  \\
C & 0.773                  & 0.370                   & 0.797                  & 0.575                    & 0.712                   & 0.581                  & 0.454                  & 0.754                    & 0.601                     & 0.887                  \\
M & 0.288                  & 0.813                   & 0.185                  & 0.951                    & 0.471                   & 0.689                  & 0.372                  & 0.467                    & 0.896                     & 0.127                  \\
R & 0.670                  & 0.424                   & 0.464                  & 0.297                    & 0.716                   & 0.745                  & 0.998                  & 0.585                    & 0.324                     & 0.380                  \\ \hline
\end{tabular}
\caption{Percentage of total species individuals occuring in each substrate according to CABOF labels. Note that a given frame may have multiple substrate labels, so a given individual may occur in multiple substrates at one time.}
\label{tab:spec_ac_sub}
\end{table*}


\section{Tasks}
\label{sec:tasks}
While our dataset has a plethora of uses, we present two specific tasks for which our dataset is well suited.

\subsection{Substrate Temporal Localization}
The first step marine researchers take to analyzing the videos that they collect is to define the temporal spans of each substrate by indicating the start and end times of each substrate as the substrate changes while the ROV drives over the ocean floor. Many substrates may occur simultaneously, which slightly complicates the problem making it a mutli-label classification problem. Our dataset makes it possible to develop and test automated methods for this problem. 

\subsection{Counting Species Individuals}
DUSIA also makes it possible to count the number of individuals of species occurring in the videos. Counting can be achieved in three stages: detection, tracking, and then counting. We present a simple baseline method for achieving this. While many computer vision methods for counting may rely on localization information such as bounding boxes, marine researchers are interested in the number of individuals occurring in the video and are less interested in where exactly in the frame an organism occurs. They can use video timestamps of those individuals' occurrence to map those timestamps back to their GPS coordinate time log from the expedition in which the video was captured, generating population density maps for different species.


Additionally, we provide bounding box labels for ten species of interest as described in Section \ref{sec:invert_annos}.

\section{Methods}
\label{sec:methods}
While our dataset can be used to train models to solve a wide variety of problems including substrate classification, species hotspot estimation, species counting, and invertebrate tracking, we present methods for substrate temporal localization and invertebrate species detection using partially supervised frames with our primary focus on invertebrate species detection. We feed our detection results to ByteTrack's tracking algorithm \cite{zhang2021bytetrack} to track invertebrate species and present a simple method for using these tracks to count invertebrate individuals.

\begin{figure*}[t]
    \centering
    \includegraphics[width=1\linewidth]{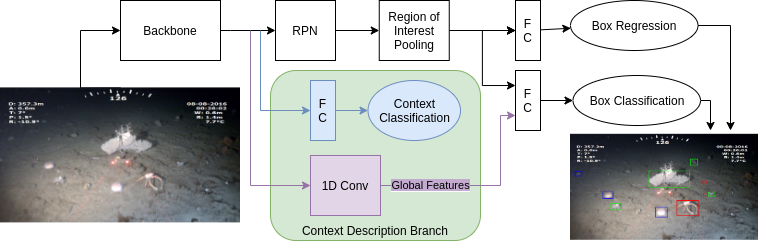}
    \caption{Context-Driven Detector: the Context Description Branch (green) takes features from the backbone, classifies context explicitly (blue), and feeds a global representation of context (purple) to the box classification layer to enhance detections. We show that using this branch enhances the detections overall indicating that learning from explicit context labels can enhance detections.}
    \label{fig:arch}
\end{figure*}

\subsection{Substrate Classification}
For a baseline, we train two basic classifiers for substrate classification. First, we trained an out-of-the-box ResNet-50 based \cite{he2016deep} classification CNN, pre-trained on ImageNet \cite{deng2009imagenet}, on frames pulled from training videos to predict four substrates at once. Then, we trained four separate ResNet-50 classifiers, one per substrate, and combined the prediction results from each of the classifiers by simply assigning each of their confidence predictions to each class since substrate classification allows multiple substrates to be present in a single frame.

\subsection{Invertebrate Species Detection}
We trained an out-of-the-box Faster RCNN model using our partially annotated keyframes (see section \ref{sec:invert_annos} for partial annotation description). We chose Faster RCNN for its adaptability and ability to classify smaller boxes, with which some object detectors struggle. As shown in Figure \ref{fig:box-dis}, many classes in DUSIA are made up of small boxes.

                                                                                                                                                                                                      
Figure \ref{fig:arch} shows vanilla Faster RCNN in black. An image is fed to a backbone network, and image features are fed to a region proposal network. Then, region of interest pooling selects proposed regions. Finally, fully connected layers classify each region and regress the bounding box coordinates to refine their localization.  We made no modifications to Faster R-CNN for this baseline model and refer to this version as vanilla Faster RCNN with the loss function, $L_{v}$, described by Ren et al \cite{ren2015faster}:
\begin{equation}
L_{v} = L_{d} + L_{p}
\label{eq:lv}
\end{equation}
where $L_{d}$ is the loss for the detector and $L_{p}$ is the loss for the region proposal network. Since we make no modifications to this part of the loss, we leave the details of the original loss description to the source paper.

\subsubsection{Negative Region Dropping}
Because much of our partially annotated training set contains unlabelled individuals of species of interest, we propose an approach for teaching the detection network to pay more attention to the true positive labels, and to pay less attention to potential false positives during training because a false positive may actually just be an unlabelled positive. There is generally no way of being sure whether an individual of interest is not present given a partially labelled training set, but all of the boxes provided for training are correct, true positive examples. Since humans can make sense of such a scenario, we aim to create a method for a detector to emulate that process.

Faster RCNN's region proposal network (RPN) generates proposals and computes a loss to learn which proposal contains an object of interest or not. Each proposal is assigned a label, positive or negative, based on whether it has sufficient overlap with a ground truth box (positive) or not (negative). Because DUSIA's training set contains unlabelled positives, we propose randomly dropping out a percentage of the negative proposals, thereby giving negative examples a lower weight and positive examples a higher weight. Dropping these negative proposals simply equates to not including them in the RPN's loss, $L_{p}$. 

We explore different percentages, $\rho$, to drop in section \ref{sec:experiments}, and show that dropping negative proposals in this way leads to significant improvement in detection performance on DUSIA. 

\subsubsection{Context Driven Detection}
To improve invertebrate detection using context annotations, we introduce the novel Context Description Branch as shown in green in Figure \ref{fig:arch}. The first iteration of the context description branch (blue in Figure \ref{fig:arch}) flattens the feature map from the backbone network and feeds this flattened vector to a fully connected layer which is trained in tandem with the detection branch to predict the multi-class substrate label. Simply backpropagating a weighted binary cross entropy loss to the backbone network to predict the substrate label increases the model's performance and generalizability (as measured by performance on the test set) by teaching the network about context via explicit context classification. This joint optimization generates cues in the backbone feature map that improve the invertebrate detection. For this iteration of the network, the loss function looks the same as equation \ref{eq:lv} with the additional loss for explicit context classification.
\begin{equation}
L = L_{v} + \alpha*L_{c}
\label{eq:l}
\end{equation}
where $\alpha$ is a hyperparameter weight and $L_{c}$ is a binary cross entropy loss for context labels.

\begin{table*}[t]
\centering
\begin{tabular}{@{}rrrrrrrr@{}}
\toprule
         & \multicolumn{1}{l}{}        & \multicolumn{1}{l}{}         & \multicolumn{4}{c}{test\_wv per class APs}                                                                                                        &                                    \\
         & \multicolumn{1}{l}{val mAP} & \multicolumn{1}{l}{test mAP} & B                                  & C                                  & M                                  & R                                  & test\_wv mAP                       \\ \midrule
Separate & \textbf{0.588}              & \textbf{0.646}               & \multicolumn{1}{r}{\textbf{0.274}} & \multicolumn{1}{r}{\textbf{0.802}} & \multicolumn{1}{r}{0.750}          & \multicolumn{1}{r}{\textbf{0.826}} & \multicolumn{1}{r}{0.663}          \\
Combined & 0.551                       & 0.572                        & \multicolumn{1}{r}{0.259}          & \multicolumn{1}{r}{0.777}          & \multicolumn{1}{r}{\textbf{0.951}} & \multicolumn{1}{r}{0.781}          & \multicolumn{1}{r}{\textbf{0.692}} \\
CDD      & 0.517                       & 0.596                        & \multicolumn{1}{r}{-}  & \multicolumn{1}{r}{-}   & \multicolumn{1}{r}{-}            & \multicolumn{1}{r}{-}   & \multicolumn{1}{r}{-}                                  \\ \bottomrule
\end{tabular}
\caption{Substrate classifier performance. Per class APs are shown for the test\_wv set. CDD shows the classification performance of the CDD with $\alpha$ = 0.0001 and $\rho$ = 0.75, which was not run on test\_wv.}
\label{tab:sub-cls}
\end{table*}

By feeding global features alongside local features to the box classification layer, we can also enhance the model's performance; however, for the network to learn from them simultaneously, the global and local features must be on similar orders of magnitude. For vanilla Faster RCNN, the local box features are vectors of size 1,024. Global features from the ResNet-50 backbone, though, are much larger. To address this size mismatch, we add a 1D convolution layer to the context description branch, which reduces the dimension of the backbone's feature map. This reduced map represents the global context information, which is largely the visible substrate, to a dimensionality on the same order of magnitude as each of the box features that are fed to the box classification head's fully-connected layer. Along those lines, we also scale the global features to match the local box feature vector by simply multiplying the global features element-wise with a scalar hyperparameter, $\beta$. 

Because Faster RCNN predicts the class of each box based on a set of box features, which is a local representation of the object that is being classified, we enhance these box classifications by concatenating each image's global context information to each of its box features. This concatenation fuses together local and global features and allows the network to draw more immediate conclusions about the global information, object features, and their relationship, which is especially relevant when classifying invertebrate species in this setting. Here, we make no changes to the loss function from equation \ref{eq:l}, and the 1D convolution kernel is learned.

\subsection{Invertebrate Tracking and Counting}
To illustrate an example pipeline for invertebrate counting, we use a detection plus tracking approach. First, we train our detector on keyframes from our training set, and then we run inference on the full validation and testing videos at 30 fps saving all detections including their spatial and temporal locations, class labels, and confidence scores.

As an intermediate step, we filter out all low confidence detections under different thresholds so that the tracker does not see low confidence detections.

ByteTrack \cite{zhang2021bytetrack} takes as input the detections (box coordinates and confidence scores) of a single class at a time and metadata from the images (e.g. image size). In short, ByteTrack performs a modified Kalman filter based algorithm to the detections in order to link them in adjacent frames and assign each detection a track ID, or filter it out. 

We apply a second filter to the output of ByteTrack such that track IDs that occur in too few frames are filtered out.

Finally, we count species individuals. To emulate the process used by marine scientists, we only count species individuals that touch the bottom of the frame. So, if a tracked species' box touches the bottom of the frame, we mark its track ID as counted and simply increment its class's count. This way, for each video, we can compute a total number of species per video that we can then compute relative error using our predicted counts and the sum of each video's CABOF labels.

\section{Experiments}
\label{sec:experiments}
We test a few models and methods for the substrate temporal localization task in an effort to provide a baseline for other works to improve upon.

\subsection{Substrate Temporal Localization}
\subsubsection{Single Classifier}
\label{sec:single}
We test a simple ResNet-50 based image classifier trained with a batch size of 32, learning rate of 0.1, and up to 50 epochs, selecting the epoch weights that perform best on the validation set. We also tested learning rates of 0.01 and 0.001 for our classifiers, and these models performed similarly but slightly worse. Table \ref{tab:sub-cls} shows the results of these experiments as predictions were made on the fully annotated frames of our validation and testing sets. These two sets are included for comparison with the context classification performance of CDD with explicit context classification, though CDD is optimized to perform detection simply using substrate prediction as a guiding sub-task. For substrate localization, though, we have annotations for almost every frame. So, we also present our classification performance on the test\_wv set which includes many more frames from the test videos. To generate test\_wv we simply sample the test videos uniformly at one frame per second. We then classify each frame, and present the AP scores.

\subsubsection{Combination of Multiple Classifiers}
As mentioned in previous sections, substrate annotations are currently completed by trained marine scientists in multiple passes through each video, one pass per substrate. Inspired by this approach, we use one binary classifier network per substrate class. Each ResNet-50 image classification network is trained independently on the training set; however, each network is trained to simply indicate whether one substrate is present or not. We use each classifier's prediction together to predict the multi-class label and refer to this method as our combined approach. Table \ref{tab:sub-cls} shows that this method improves performance over a single multi-classifier for most substrates, indicating that each approach may have different use cases.

All classifiers seem to struggle with correctly identifying the boulder substrate, and, given the nuance in differences between hard substrates, this is not surprising considering the classifiers have little scale information to use to determine and differentiate exact sizes of different pieces of cobble, boulders, or larger rock formations. Additionally, the changing perspective of the ROV makes it difficult to understand scale in the videos. That said, a dedicated boulder detector out-performed the single classifier method overall due to its impressive performance classifying the mud class.


\subsection{Invertebrate Species Detection}
In order to detect species individuals, we present mean average precision (mAP) results for object detection with an intersection over union (IOU) threshold of 0.5. For each detection experiment, we initalize our models with weights pretrained on ImageNet and then train the network for up to 15 epochs. We select the model from the epoch with the best performance on the fully annotated frames of the validation set. Then, we run inference on the fully annotated frames of the test set using those selected model weights.  We repeat the training and testing procedure four times for each experiment and report the average results over the four runs because PyTorch does not support deterministic training for our model at the time of writing.

We first train vanilla Faster RCNN \cite{ren2015faster} with a batch size of 8 and try several learning rates after initializing with weights pre-trained on COCO \cite{lin2014microsoft} provided by PyTorch \cite{paszke2019pytorch}. The results are shown in Table \ref{tab:van-lr}.

\begin{table}[t]
\centering
\begin{tabular}{rrr}
\hline
\multicolumn{1}{l}{lr} & \multicolumn{1}{l}{val mAP} & \multicolumn{1}{l}{test mAP} \\ \hline
0.1                    & 0.454                       & 0.361                        \\
0.01                   & \textbf{0.490}              & \textbf{0.391}               \\
0.001                  & 0.482                       & 0.367                        \\ \hline
\end{tabular}
\caption{Performance of vanilla Faster RCNN with varying learning rates}
\label{tab:van-lr}
\end{table}

\begin{table}[t]
\centering
\begin{tabular}{rrrr}
\hline
\multicolumn{1}{l}{lr} & \multicolumn{1}{l}{$\rho$} & \multicolumn{1}{l}{val mAP} & \multicolumn{1}{l}{test mAP} \\ \hline
0.01                   & 0                          & 0.490                       & 0.391                        \\
0.01                   & 0.5                        & 0.492                       & 0.413                        \\
0.01                   & 0.75                       & \textbf{0.509}              & \textbf{0.439}               \\
0.01                   & 0.9                        & 0.492                       & 0.403                        \\
0.01                   & 1                          & 0.297                       & 0.264                        \\
0.001                  & 0.75                       & 0.479                       & 0.380                        \\
0.001                  & 0.9                        & 0.481                       & 0.380                        \\ \hline
\end{tabular}
\caption{Performance of Faster-RCNN with varying Negative Region Dropping percentages}
\label{tab:rho-only}
\end{table}

We then perform hyperparameter searches for each of our method contributions described in section \ref{sec:methods}: $\alpha$ for explicit context learning and backbone refinement, $\beta$ for global context feature fusion, and $\rho$ for Negative Region Dropping. After testing each hyperparameter independently, we try combinations of each and discuss the results.  We prioritize test mAP over val mAP as test mAP is more indicative of the generalizability of our model since the best model weights are selected on best val mAP.

\subsubsection{Negative Region Dropping Percent \texorpdfstring{$\rho$}{Lg}}
Table \ref{tab:rho-only} shows that Negative Region Dropping consistently improves the training on DUSIA by teaching the network to focus more on learning from true examples than negative examples. Interestingly, setting $\rho$ to 1.0 detrimentally harms performance indicating that having some negative regions contribute to the region proposal loss is still important.

\subsubsection{Global Feature Fusion Scalar \texorpdfstring{$\beta$}{Lg}}
By creating a global feature representation and feeding it later in the network, the network is better able to classify boxes correctly, but concatenating a global feature representation with the local box features requires that the features come in at similar scales. Table \ref{tab:beta-only} shows the effect of different scalar values for this fusion.

\begin{table}[t]
\centering
\begin{tabular}{rrrr}
\hline
\multicolumn{1}{l}{lr} & \multicolumn{1}{l}{$\beta$} & \multicolumn{1}{l}{val mAP} & \multicolumn{1}{l}{test mAP} \\ \hline
0.01                   & 0                           & 0.490                       & 0.391                        \\
0.01                   & 0.1                         & 0.471                       & 0.371                        \\
0.01                   & 0.01                        & 0.491                       & 0.397                        \\
0.01                   & 0.001                       & \textbf{0.499}              & 0.396                        \\
0.01                   & 0.0001                      & 0.494                       & \textbf{0.410}               \\
0.01                   & 1.0E-05                     & 0.496                       & 0.406                        \\
0.01                   & 1.0E-06                     & 0.482                       & 0.394                        \\
0.001                  & 0.01                        & 0.475                       & 0.374                        \\
0.001                  & 0.001                       & 0.477                       & 0.371                        \\ \hline
\end{tabular}
\caption{Performance of the Context Driven Detector given different $\beta$ scalar values}
\label{tab:beta-only}
\end{table}

\subsubsection{Context Loss Weight \texorpdfstring{$\alpha$}{Lg}}
By modifying the detector to simultaneously classify the context of an image in parallel with detection, we demonstrate that simply backpropagating information useful for classifying substrate to the backbone also serves to help improve detection performance. Training a joint task in this way leads to less powerful context classifications than a dedicated context classifier, but it leads to a more powerful object detector. Table \ref{tab:alpha-only} shows the effects of $\alpha$ on the detection performance.

\begin{table}[t]
\centering
\begin{tabular}{rrrr}
\hline
\multicolumn{1}{l}{lr} & \multicolumn{1}{l}{$\alpha$} & \multicolumn{1}{l}{val mAP} & \multicolumn{1}{l}{test mAP} \\ \hline
0.01                   & 0                            & 0.490                       & 0.391                        \\
0.01                   & 0.1                          & 0.470                       & 0.389                        \\
0.01                   & 0.01                         & 0.494                       & 0.419                        \\
0.01                   & 0.001                        & 0.487                       & 0.401                        \\
0.01                   & 0.0001                       & 0.502                       & \textbf{0.420}               \\
0.01                   & 1.0E-05                      & \textbf{0.507}              & 0.410                        \\
0.01                   & 1.0E-06                      & 0.501                       & 0.408                        \\
0.001                  & 0.01                         & 0.456                       & 0.358                        \\
0.001                  & 0.001                        & 0.453                       & 0.361                        \\ \hline
\end{tabular}
\caption{Performance of the Context Driven Detector given different context loss scaling $\alpha$ values}
\label{tab:alpha-only}
\end{table}

\subsubsection{Hyperparameter Combinations}
We illustrate that each hyperparameter alone can improve the detector performance over the baseline out-of-the-box models. We further illustrate that Negative Region Dropping and context driven detection can work in tandem to further improve performance. We also find that a context driven detector with both implicit attention to context (global feature fusion) and explicit context classification does not necessarily outperform implicit context usage or explicit classification only. Training on both implicit and explicit context simultaneously may interfere with each other. Still, we emphasize that learning from context can significantly improve object detection performance in this setting, and we aim to find even better ways to utilize contextual information to better classify objects in future work. 

Table \ref{tab:best-models} highlights the best hyperparameter settings revealed during our search, and Appendix \ref{sec:append} goes into more detail on the settings tested for this study. Note that the $\beta$ column set to zero indicates that global features are not being scaled by 0, rather they are not being concatenated with the local box features at all.

We find that Negative Region Dropping increases the overall performance of both vanilla Faster RCNN and context driven detectors. While explicit and implicit context usage may conflict with one another in training, independently they can achieve performance increases. The best model overall is achieved with global context feature fusion and Negative Region Dropping, and a model with explicit context classification and Negative Region Dropping follows close behind. We find that using context to influence detections leads to a 7.4\% increase, using negative region dropping leads to a 12.3\%, and together they can achieve a 14.3\% increase in mAP on the fully annotated frames in DUSIA's test set.

\begin{figure}[t]
    \centering
    \includegraphics[width=1\linewidth]{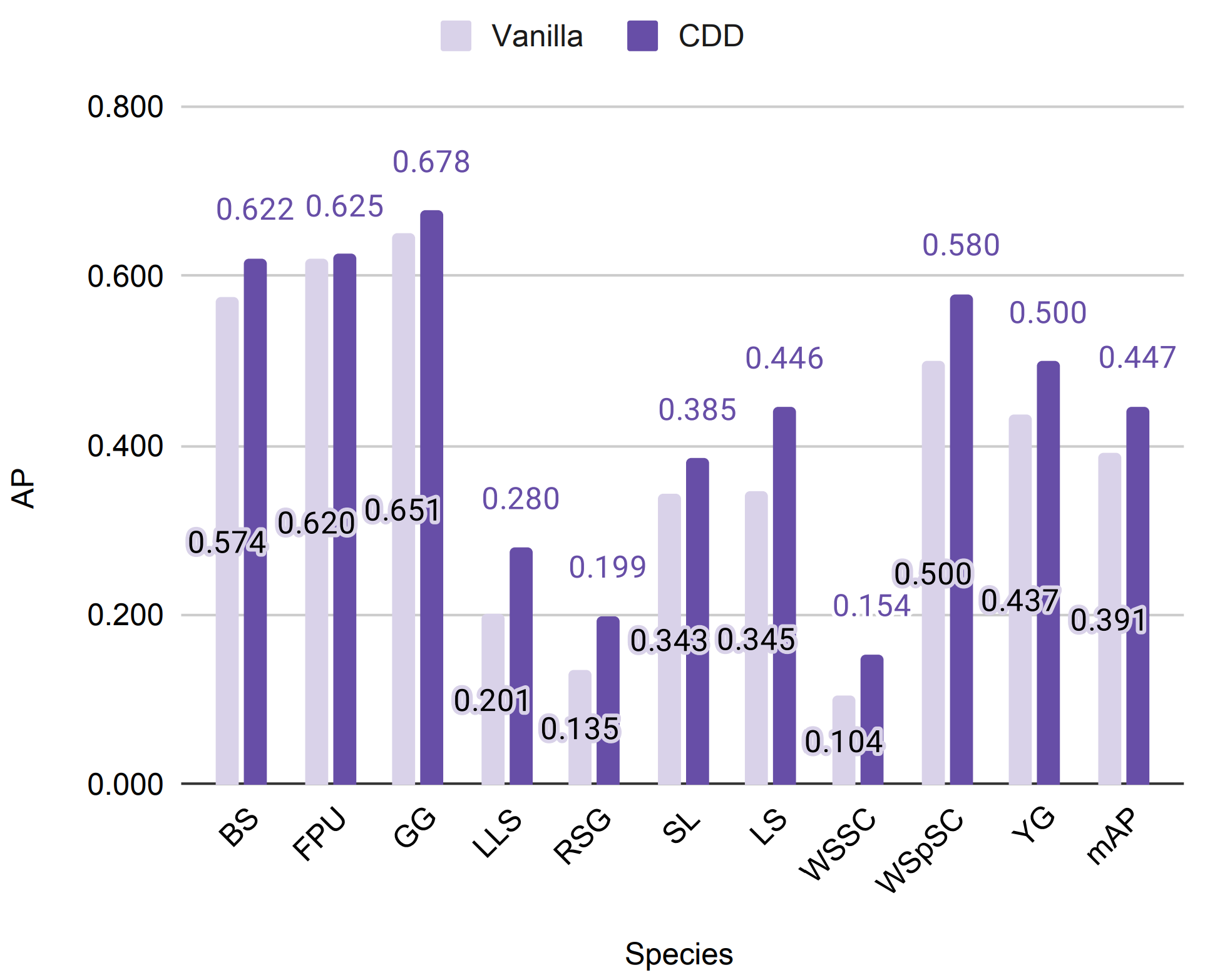}
    \caption{Per class test AP comparison of vanilla Faster RCNN and the best Context Driven Detector}
    \label{fig:per-class}
\end{figure}

Figure \ref{fig:per-class} illustrates the per class AP detection performance of our best model compared with vanilla Faster RCNN showing that our model significantly increases performance on all classes. Figure \ref{fig:dets} shows qualitative examples of success and failure cases of the best version of CDD.

\begin{table}[t]
\centering
\begin{tabular}{rrrrr}
\hline
\multicolumn{1}{l}{$\alpha$} & \multicolumn{1}{l}{$\beta$} & \multicolumn{1}{l}{$\rho$} & \multicolumn{1}{l}{val mAP} & \multicolumn{1}{l}{test mAP} \\ \hline
0                            & 0                           & 0                          & 0.490                       & 0.391                        \\
0                            & 0.0001                      & 0                          & 0.494                       & 0.410                        \\
0.01                         & 0.1                         & 0                          & 0.480                       & 0.420                        \\
0.0001                       & 0                           & 0                          & 0.502                       & 0.420                        \\
1.0E-06                      & 0.01                        & 0.75                       & 0.517                       & 0.430                        \\
0                            & 0                           & 0.75                       & 0.509                       & 0.439                        \\
0.0001                       & 0                           & 0.75                       & 0.514                       & 0.439                        \\
0                            & 0.01                        & 0.75                       & \textbf{0.524}              & \textbf{0.447}               \\ \hline
\end{tabular}
\caption{Performance of best models from each hyperparameter combination}
\label{tab:best-models}
\end{table}

\def \supfih {2cm}
\begin{figure*}[t!]
    \centering
    \includegraphics[height=\supfih{}]{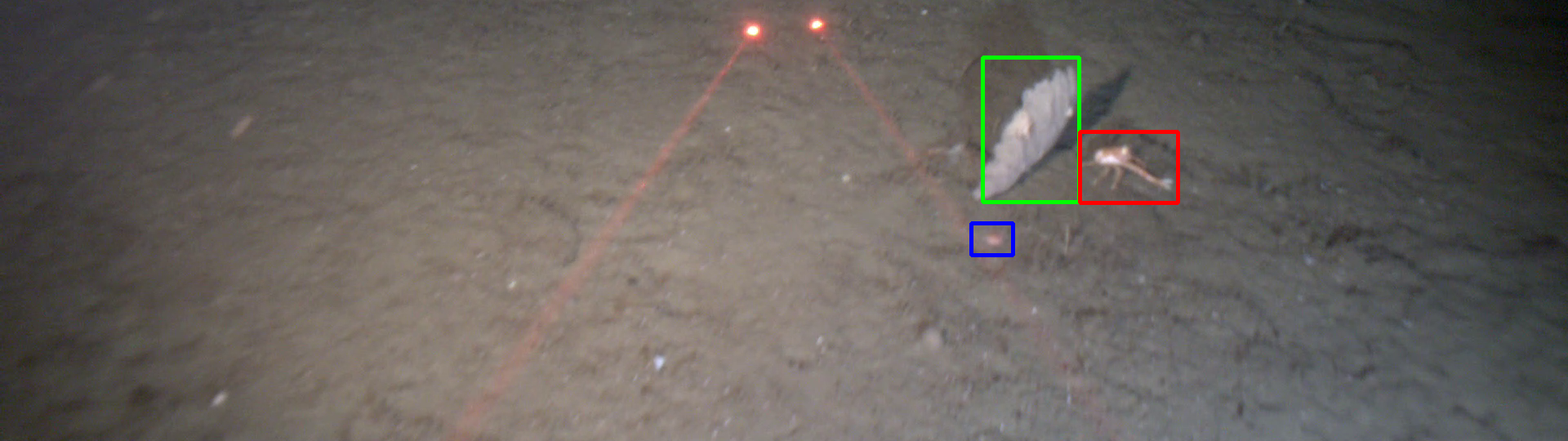}
    \includegraphics[height=\supfih{}]{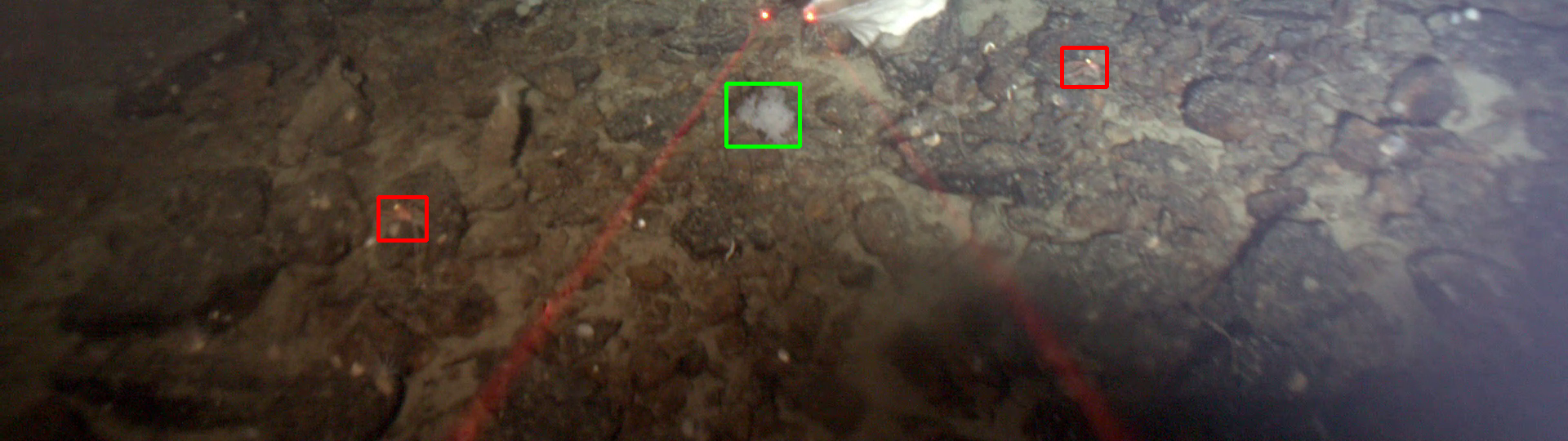}
    \includegraphics[height=\supfih{}]{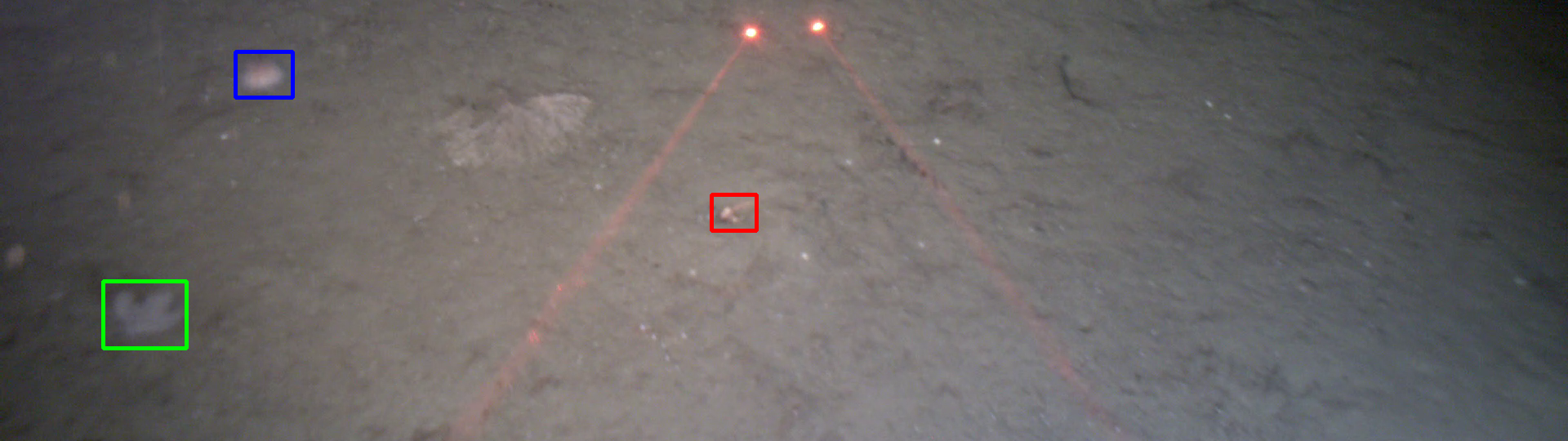}
    \includegraphics[height=\supfih{}]{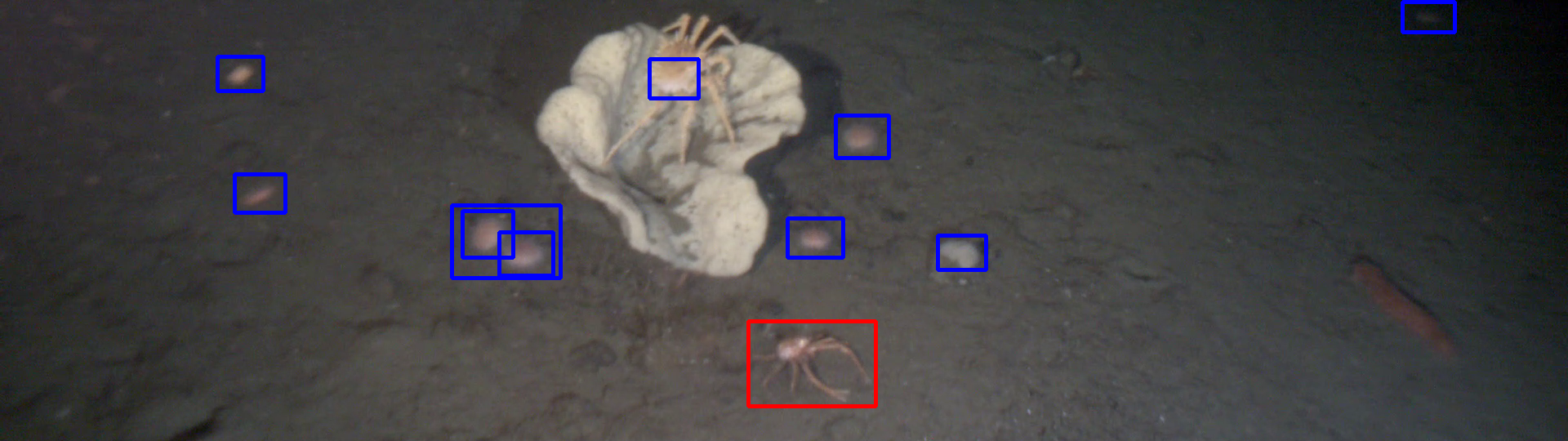}
\caption{Detection examples from our dataset. Blue indicates fragile pink urchin; green, gray gorgonian; and red, squat lobster. We show the success of our detector with the exception of the bottom right image. A crab (not a species of interest) is mislabeled as a fragile pink urchin toward the top center of the image. In the left side of the image, two pieces of floating debris are labelled as urchins, and close to the center two urchins are counted thrice. Right of center, a rock is labelled as an urchin. These failure cases demonstrate some of the challenges of DUSIA. In the top right corner of the bottom right image, a very difficult to see pink urchin is correctly detected.}
\label{fig:dets}
\end{figure*}

\subsection{Invertebrate Species Counting}
There are some noteworthy differences between the detection and counting problems. As mentioned in Section \ref{sec:splits}, we partition DUSIA's videos into three sets: training, validation, and testing sets. However, the detector sees only a small fraction of each video as only a small subset of each video has bounding box annotations. Further, while we refer to three of our videos as validation videos, our detection models do not train on those videos at all, and only 514 frames from those ~124,000 validation video frames are used in the detection validation process to select our best model weights.

\begin{table*}[!t]\centering
\small
\scriptsize
\resizebox{0.9\linewidth}{!}{
\begin{tabular}{rrrrrrrrrrrrr}\toprule
\multicolumn{12}{c}{val set per species relative errors} & \\\cmidrule{1-13}
$\gamma$ & $\tau$ & BS &FPU &GG &LLS &RSG &SL &LS &WSSC &WSpSC &YG &mean \\
0 & 0 & 11.2 &4.04 &5.75 &25.6 &60.9 &3.18 &\cellcolor[HTML]{e5eef0}0.35 &2.98 &2.32 &18.7 &13.5 \\
20 & 0.5 & \cellcolor[HTML]{dce8ea}-0.18 &\cellcolor[HTML]{d6e4e7}-0.091 &\cellcolor[HTML]{e6eef0}-0.34 &1.13 &\cellcolor[HTML]{d7e5e7}-0.11 &\cellcolor[HTML]{f0f5f6}-0.50 &-0.90 &-0.88 &\cellcolor[HTML]{e2ecee}-0.27 &\cellcolor[HTML]{d0e0e3}0.00 &\cellcolor[HTML]{ebf2f3}0.439 \\

\hline\\

\multicolumn{12}{c}{test set per species relative errors}\\\cmidrule{1-13}
$\gamma$ & $\tau$ &BS &FPU &GG &LLS &RSG &SL &LS &WSSC &WSpSC &YG &mean \\
0 & 0 & 6.00 &4.73 &15.38 &46.66 &70.23 &3.29 &2.57 &2.84 &3.71 &12.21 &16.8 \\
20 & 0.5 & \cellcolor[HTML]{f3f8f8}-0.56 &\cellcolor[HTML]{d8e5e8}0.14 &\cellcolor[HTML]{d2e2e4}-0.03 &1.28 &\cellcolor[HTML]{e0ebed}-0.25 &\cellcolor[HTML]{f1f6f7}-0.51 &-0.84 &-0.91 &\cellcolor[HTML]{dfeaec}-0.24 &\cellcolor[HTML]{e9f1f2}-0.39 &\cellcolor[HTML]{f0f5f6}0.515 \\
\bottomrule
\end{tabular}
}

\caption{Relative errors of our counting method with no thresholding and the best threshold settings. Darker color indicates better performance. See Table \ref{tab:count_ac_sp} for ground truth counts for each species.}
\label{tab:err}
\end{table*}

In contrast, our counting method runs our detector on the entire lengths of the videos in the validation and testing sets, posing a great challenge to the generalizability and robustness of an object detection model. That is, the sets of frames used for the counting task are much larger than those used for detection. Also, the frames annotated with invertebrate species (i.e. all the frames in the detector's training set) all include instances of those species of interest. In contrast, each video contains long time spans of both densely and sparsely annotated areas including some long regions with no species of interest. As a result, counting species individuals poses a very challenging problem, and much work remains to be done in the power of a detector and its ability to differentiate between background and species of interest in both sparsely and densely populated environments.

Still, we aim to demonstrate the challenge of this problem with a simple baseline method, though much work remains to be done to achieve a result that would be able to replace the annotation abilities of trained marine scientists. We hope that DUSIA can aid in pushing the limits of computer vision models and extend computer vision methods' usefulness into more challenging, scientific data.

In order to count invertebrate individuals, we first run the best performing version of CDD on each of our val and test videos at the full frame rate of 30 fps and save all detections. Then, we filter out all detections with confidence scores under a threshold, $\tau$, before feeding all detections to ByteTracker. We then filter the output of ByteTrack by discarding any track IDs with less than $\gamma$ detections in the track. That is, if a track ID is assigned to boxes in only a few frames, we discard that track ID. We experimented with ByteTracker's hyperparameters and found that their effect was significantly smaller than the effects of $\tau$ and $\gamma$, so we opt to use the default hyperparameter settings for ByteTracker. We leave the details of ByteTracker to the original work \cite{zhang2021bytetrack}. Finally, for each species, we count the number of that species' track IDs that touch the bottom of any frame.

We applied the two aforementioned filters because, without any filters, our method vastly over counts all species through all videos. Figure \ref{fig:dets} shows examples of a few false positive detections, and these types of errors likely contribute heavily to our method's over counting as the detector is run over hours of videos, accumulating false positive results. 

To address the over counting issue, we opted to feed the tracker only our most confident detections and to only count tracks that occur across multiple frames. This filtering significantly improved the performance, but the error remains unacceptably high.

Table \ref{tab:err} shows the relative error for each class on the val and test videos as well as the mean relative error, averaged over all classes, as we vary the $\tau$ and $\gamma$ parameters. We leave the error sign to indicate over (positive error) or under (negative error) counting, but we compute the mean errors using the absolute value of the error values for each class. Clearly, the detector hardly learns some of the rarer classes (e.g. long-legged sunflower star and red swiftia gorgonian) and regularly misclassifies background, which may include species outside of our ten species of interest, as our species of interest. Appendix \ref{sec:append} contains more experiment error results for varying these filter thresholds.

Ultimately, these baseline results indicate that this simple method is not powerful enough to put into practice given the effectiveness of our current detection model. Much work on methods for this problem is left to be done. We could look deeper into per class thresholds, but we expect improving object detections, false positive filtering, and the tracking algorithm would be more robust. We leave these improvements to future work.

\section{Discussion and Future Work}
\label{sec:discussion}
Our baseline methods' detection and counting performance leaves plenty of room for improvement. Our detection methods do not enforce any sort of temporal continuity present in the ROV videos, which could likely improve performance, and the methods do not yet take advantage of the abundant, weak CABOF labels during training. 

It is interesting to find the difference in performance of the different types of substrate classifiers. Overall, the substrate classification results are good enough for some substrates, and in future work we hope to see results good enough to fully automate this process. Additionally, marine scientists are interested in real time substrate classifiers that can indicate which substrates the ROV is passing in real time. Any indication of species hotspots in real time during expeditions can improve each excursion's productivity by reducing more manual means of searching for given substrates, habitats, and species hotspots. 

The detection results of the Context Driven Detector provide a baseline, but in order to fully translate these detections to tracks with individual re-identification and counting, there is much work to be done. We hope to next take full advantage of the CABOF labels and to use context in more powerful ways to improve detection performance in future work. Further, we plan to enforce temporal continuity to improve our counting predictions. These improvements can lead us to eventually begin automating some of the invertebrate counting that is currently done manually. 

By making DUSIA public, we also invite other collaborators to work independently or in cooperation with us to help improve our methods. 

\backmatter

\section*{Supplementary information}
DUSIA's data, annotations, and baseline methods will be made publicly available at the time of publication. 

\section*{Acknowledgments}
This research was supported in part by National Science Foundation (NSF) award: SSI \# 1664172. We would like to thank Dirk Rosen and Andy Lauermann from Marine Applied Rsearch \& Exploration group for their video collection, guidance, and help through this project. We would also like to thank Anmol Kapoor and Shafin Haque for their contributions to the project.

\section*{Statements and Declarations}
\begin{itemize}
\item Funding: this work was partially supported by National Science Foundation (NSF) award: SSI \# 1664172
\item Conflict of interest/Competing interests: n/a
\item Ethics approval : n/a
\item Consent to participate: n/a
\item Consent for publication: n/a
\item Availability of data and materials: data and annotations will be made publicly available at the time of publication.
\item Code availability: code and implementation of methods will be made publicly available at the time of publication.
\end{itemize}

\newpage
\clearpage
\bibliographystyle{apacite}
\bibliography{sn-bibliography}

\newpage
\clearpage
\begin{appendices}

\onecolumn
\section{Species Statistics}
\label{sec:all-spec}

\begin{table*}[ht!]
\label{tab:all-spec}
\centering
\resizebox{0.8\linewidth}{!}{
\begin{tabular}{@{}lr|lr@{}}
\toprule
Species                      & \multicolumn{1}{l|}{Count} & Species                   & \multicolumn{1}{l}{Count} \\ \midrule
\textbf{Fragile pink urchin} & \textbf{3,402}              & Spot prawn                & 18                        \\
\textbf{Squat lobster}       & \textbf{2,593}              & UI anemone                & 17                        \\
UI lobed sponge              & 1,753                       & Thorny sea star           & 16                        \\
\textbf{White slipper sea cucumber}   & \textbf{1,313}                       & UI anemone 2              & 14                        \\
\textbf{Laced sponge}              & \textbf{632}                       & California king crab      & 13                        \\
\textbf{Gray gorgonian}      & \textbf{556}               & UI trumpet sponge         & 12                        \\
UI hairy boot sponge         & 426                        & Pom-pom anemone           & 11                        \\
\textbf{Basket star}                & \textbf{361}                        & UI prawn                  & 9                         \\
\textbf{White spine sea cucumber}     & \textbf{318}                       & Crested sea star          & 8                         \\
\textbf{Long legged sunflower star}   & \textbf{306}                       & White sea pen             & 8                         \\
\textbf{Red swiftia gorgonian}        & \textbf{257}                        & Red sea star              & 8                         \\
UI branched sponge           & 228                        & UI sea pen                & 7                         \\
UI vase sponge               & 210                        & Solaster sun star complex & 6                         \\
\textbf{Yellow gorgonian}           & \textbf{150}                        & UI octopus                & 5                         \\
UI boot sponge               & 128                        & UI nipple sponge          & 4                         \\
Cookie star                  & 90                         & UI gorgonian              & 3                         \\
UI anemone 4                 & 67                         & Spiny/thorny star complex & 3                         \\
UI sea star                  & 54                         & Gray moon sponge          & 2                         \\
UI tubeworm                  & 50                         & Brown box crab            & 2                         \\
Henricia complex             & 47                         & Decorator crab            & 2                         \\
UI large yellow sponge       & 44                         & UI sand dwelling anemone  & 2                         \\
UI thin red star             & 39                         & UI nudibranch             & 1                         \\
UI orange gorgonian          & 38                         & Orange puffball sponge    & 1                         \\
Mushroom soft coral          & 36                         & Red octopus               & 1                         \\
Black coral                  & 34                         & Red gorgonian             & 1                         \\
Benthic siphonophore         & 25                         & Rose star                 & 1                         \\
Bubblegum coral              & 24                         & Cushion star              & 1                         \\
Deep sea cucumber            & 20                         & UI urchin                 & 1                         \\
Fish eating star             & 19                         & UI anemone 1              & 1                         \\
Spiny red star               & 18                         &                           & \multicolumn{1}{l}{}      \\ \bottomrule
\end{tabular}
}
\caption{All species and their counts in DUSIA. Bold shows the species that also include bounding box annotations. UI stands for unidentified and is used when organism's exact species cannot be determined.}
\end{table*}

\begin{figure*}[t!]
    \centering
    \includegraphics[width=0.75\linewidth]{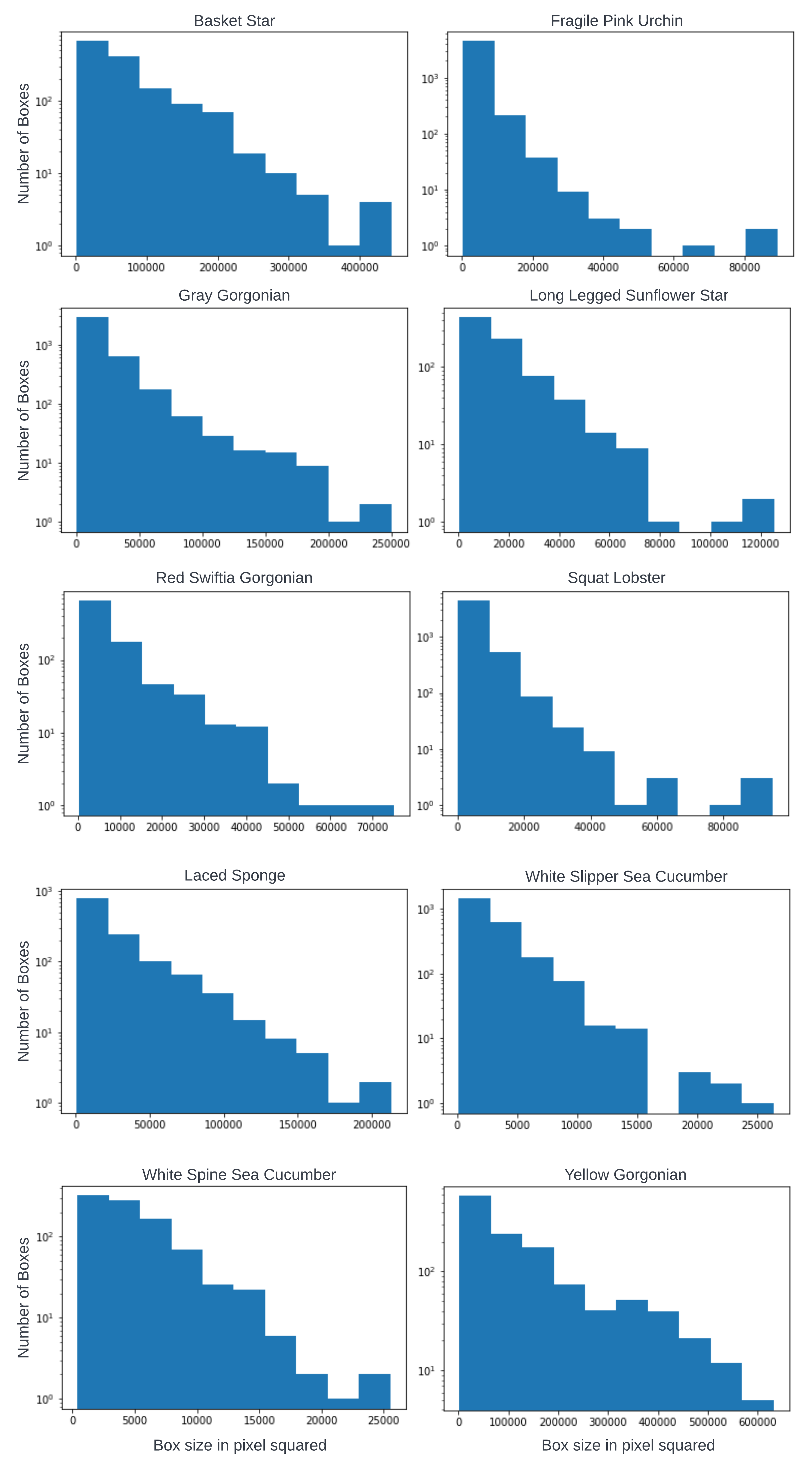}
    \caption{Histograms illustrating the distributions of box sizes (in pixels squared) for each species of interest.}
    \label{fig:box-dis}
\end{figure*} 

\newpage
\clearpage
\onecolumn
\section{Hyperparameter Search Summary}
\label{sec:append}

\begin{table}[h!]
\centering
\resizebox{\linewidth}{!}{
\begin{tabular}{@{}rrrrrr|rrrrrr@{}}
\toprule
\multicolumn{1}{l}{lr} & \multicolumn{1}{l}{$\alpha$} & \multicolumn{1}{l}{$\beta$} & \multicolumn{1}{l}{$\rho$} & \multicolumn{1}{l}{val mAP} & \multicolumn{1}{l|}{test mAP} & \multicolumn{1}{l}{lr} & \multicolumn{1}{l}{$\alpha$} & \multicolumn{1}{l}{$\beta$} & \multicolumn{1}{l}{$\rho$} & \multicolumn{1}{l}{val mAP} & \multicolumn{1}{l}{test mAP} \\ \midrule
0.1                    & 0                            & 0                           & 0                          & 0.454                       & 0.361                         & 0.01                   & 1.00E-04                     & 0.01                        & 0                          & 0.487                       & 0.405                        \\
0.01                   & 0                            & 0                           & 0                          & 0.490                       & 0.391                         & 0.01                   & 1.00E-05                     & 0.01                        & 0                          & 0.489                       & 0.404                        \\
0.001                  & 0                            & 0                           & 0                          & 0.482                       & 0.367                         & 0.01                   & 1.00E-06                     & 1.00E-02                    & 0                          & 0.486                       & 0.404                        \\
0.01                   & 0.1                          & 0                           & 0                          & 0.470                       & 0.389                         & 0.01                   & 1.00E-04                     & 1.00E-04                    & 0                          & 0.471                       & 0.395                        \\
0.01                   & 0.01                         & 0                           & 0                          & 0.494                       & 0.419                         & 0.01                   & 1.00E-05                     & 1.00E-04                    & 0                          & 0.487                       & 0.383                        \\
0.01                   & 1.00E-03                     & 0                           & 0                          & 0.487                       & 0.401                         & 0.01                   & 1.00E-04                     & 0.001                       & 0                          & 0.491                       & 0.388                        \\
0.01                   & 1.00E-04                     & 0                           & 0                          & 0.502                       & 0.420                         & 0.001                  & 0.01                         & 0.001                       & 0                          & 0.469                       & 0.373                        \\
0.01                   & 1.00E-05                     & 0                           & 0                          & 0.507                       & 0.410                         & 0.001                  & 0.001                        & 0.001                       & 0                          & 0.477                       & 0.377                        \\
0.01                   & 1.00E-06                     & 0                           & 0                          & 0.501                       & 0.408                         & 0.01                   & 0.01                         & 0                           & 0.75                       & 0.491                       & 0.405                        \\
0.001                  & 0.01                         & 0                           & 0                          & 0.456                       & 0.358                         & 0.01                   & 0.001                        & 0                           & 0.75                       & 0.487                       & 0.406                        \\
0.001                  & 0.001                        & 0                           & 0                          & 0.453                       & 0.361                         & 0.01                   & 1.00E-04                     & 0                           & 0.75                       & 0.514                       & 0.433                        \\
0.01                   & 0                            & 0.1                         & 0                          & 0.471                       & 0.371                         & 0.01                   & 1.00E-05                     & 0                           & 0.75                       & 0.503                       & 0.417                        \\
0.01                   & 0                            & 0.01                        & 0                          & 0.491                       & 0.397                         & 0.01                   & 1.00E-06                     & 0                           & 0.75                       & 0.503                       & 0.433                        \\
0.01                   & 0                            & 1.00E-03                    & 0                          & 0.499                       & 0.396                         & 0.01                   & 0.1                          & 0                           & 0.9                        & 0.500                       & 0.431                        \\
0.01                   & 0                            & 1.00E-04                    & 0                          & 0.494                       & 0.410                         & 0.01                   & 0.01                         & 0                           & 0.9                        & 0.504                       & 0.421                        \\
0.01                   & 0                            & 1.00E-05                    & 0                          & 0.482                       & 0.395                         & 0.01                   & 0                            & 0.1                         & 0.75                       & 0.512                       & 0.435                        \\
0.01                   & 0                            & 1.00E-06                    & 0                          & 0.482                       & 0.394                         & 0.01                   & 0                            & 0.01                        & 0.75                       & \textbf{0.524}              & \textbf{0.447}               \\
0.001                  & 0                            & 0.01                        & 0                          & 0.475                       & 0.374                         & 0.01                   & 0                            & 1.00E-03                    & 0.75                       & 0.513                       & 0.426                        \\
0.001                  & 0                            & 0.001                       & 0                          & 0.477                       & 0.371                         & 0.01                   & 0                            & 1.00E-04                    & 0.75                       & 0.506                       & 0.420                        \\
0.1                    & 0                            & 0                           & 0.75                       & 0.456                       & 0.354                         & 0.01                   & 0                            & 1.00E-05                    & 0.75                       & 0.506                       & 0.436                        \\
0.01                   & 0                            & 0                           & 0.25                       & 0.485                       & 0.392                         & 0.01                   & 0                            & 0.01                        & 0.9                        & 0.497                       & 0.402                        \\
0.01                   & 0                            & 0                           & 0.5                        & 0.492                       & 0.413                         & 0.01                   & 0                            & 0.001                       & 0.9                        & 0.512                       & 0.430                        \\
0.01                   & 0                            & 0                           & 0.75                       & 0.509                       & 0.439                         & 0.01                   & 0.01                         & 1.00E-02                    & 0.75                       & 0.503                       & 0.412                        \\
0.01                   & 0                            & 0                           & 1                          & 0.297                       & 0.264                         & 0.01                   & 0.01                         & 1.00E-01                    & 0.75                       & 0.502                       & 0.414                        \\
0.01                   & 0                            & 0                           & 0.9                        & 0.492                       & 0.403                         & 0.01                   & 0.1                          & 0.01                        & 0.75                       & 0.515                       & 0.427                        \\
0.001                  & 0                            & 0                           & 0.75                       & 0.479                       & 0.380                         & 0.01                   & 0.1                          & 0.1                         & 0.75                       & 0.513                       & 0.437                        \\
0.001                  & 0                            & 0                           & 0.9                        & 0.481                       & 0.380                         & 0.01                   & 0.01                         & 0.001                       & 0.75                       & 0.516                       & 0.428                        \\
0.1                    & 0.1                          & 0.1                         & 0                          & 0.451                       & 0.372                         & 0.01                   & 0.1                          & 0.001                       & 0.75                       & 0.510                       & 0.419                        \\
0.1                    & 0.01                         & 0.1                         & 0                          & 0.462                       & 0.371                         & 0.01                   & 0.01                         & 0.01                        & 0.9                        & 0.508                       & 0.420                        \\
0.1                    & 0.01                         & 0.01                        & 0                          & 0.454                       & 0.375                         & 0.01                   & 0.01                         & 0.001                       & 0.9                        & 0.497                       & 0.418                        \\
0.01                   & 0.1                          & 0.1                         & 0                          & 0.450                       & 0.370                         & 0.01                   & 0.1                          & 0.001                       & 0.9                        & 0.503                       & 0.417                        \\
0.01                   & 0.01                         & 0.1                         & 0                          & 0.480                       & 0.420                         & 0.01                   & 0.001                        & 0.01                        & 0.75                       & 0.509                       & 0.427                        \\
0.01                   & 0.1                          & 0.01                        & 0                          & 0.497                       & 0.399                         & 0.01                   & 1.00E-04                     & 0.01                        & 0.75                       & 0.510                       & 0.425                        \\
0.01                   & 0.01                         & 0.01                        & 0                          & 0.489                       & 0.403                         & 0.01                   & 1.00E-04                     & 1.00E-04                    & 0.75                       & 0.515                       & 0.433                        \\
0.01                   & 0.01                         & 0.001                       & 0                          & 0.488                       & 0.408                         & 0.01                   & 1.00E-05                     & 0.01                        & 0.75                       & 0.509                       & 0.428                        \\
0.01                   & 0.001                        & 0.01                        & 0                          & 0.486                       & 0.396                         & 0.01                   & 1.00E-06                     & 0.01                        & 0.75                       & 0.517                       & 0.430                        \\
0.01                   & 0.001                        & 0.001                       & 0                          & 0.492                       & 0.396                         & \multicolumn{1}{l}{}   & \multicolumn{1}{l}{}         & \multicolumn{1}{l}{}        & \multicolumn{1}{l}{}       & \multicolumn{1}{l}{}        & \multicolumn{1}{l}{}         \\ \bottomrule
\end{tabular}
}
\caption{Results of hyperparameter search experiments on learning rate, $\alpha$, $\beta$, and $\rho$}
\label{tab:apb-hyps}
\end{table}


\begin{table*}[!t]\centering
\scriptsize
\resizebox{0.9\linewidth}{!}{
\begin{tabular}{rrrrrrrrrrrrrr}\toprule
\multicolumn{13}{c}{val set per species errors} \\\cmidrule{1-13}
\textbf{$\gamma$} &\textbf{$\tau$} &BS &FPU &GG &LLS &RSG &SL &LS &WSSC &WSpSC &YG &mean \\
0 &0 &11.24 &4.04 &5.75 &25.63 &60.89 &3.18 &\cellcolor[HTML]{e5eef0}0.35 &2.98 &2.32 &18.7 &13.5 \\
10 &0 &1.65 &\cellcolor[HTML]{d2e1e4}0.05 &\cellcolor[HTML]{d0e0e3}0.01 &2.50 &3.00 &\cellcolor[HTML]{e1ebed}-0.26 &\cellcolor[HTML]{fcfdfe}-0.70 &-0.74 &\cellcolor[HTML]{d9e6e9}-0.14 &1.89 &1.09 \\
15 &0 &0.76 &\cellcolor[HTML]{d5e3e6}-0.06 &\cellcolor[HTML]{d9e6e9}-0.14 &1.25 &\cellcolor[HTML]{fafcfc}0.68 &\cellcolor[HTML]{eaf1f3}-0.41 &-0.80 &-0.80 &\cellcolor[HTML]{e2ecee}-0.27 &1.22 &\cellcolor[HTML]{f8fafa}0.641 \\
18 &0 &\cellcolor[HTML]{f1f5f6}0.53 &\cellcolor[HTML]{d6e4e7}-0.09 &\cellcolor[HTML]{dbe8ea}-0.17 &1.00 &\cellcolor[HTML]{e7eff0}0.37 &\cellcolor[HTML]{edf3f4}-0.45 &-0.82 &-0.84 &\cellcolor[HTML]{e4eeef}-0.32 &1.00 &\cellcolor[HTML]{f3f7f7}0.560 \\
20 &0 &\cellcolor[HTML]{f1f5f6}0.53 &\cellcolor[HTML]{d6e4e7}-0.09 &\cellcolor[HTML]{dde9eb}-0.20 &1.00 &\cellcolor[HTML]{d6e4e6}0.11 &\cellcolor[HTML]{eef4f5}-0.47 &-0.85 &-0.86 &\cellcolor[HTML]{e4eeef}-0.32 &0.89 &\cellcolor[HTML]{f1f5f6}0.531 \\
22 &0 &\cellcolor[HTML]{e9f1f2}0.41 &\cellcolor[HTML]{d7e5e7}-0.10 &\cellcolor[HTML]{e0ebed}-0.25 &1.00 &\cellcolor[HTML]{d4e3e5}-0.05 &\cellcolor[HTML]{f1f6f7}-0.51 &-0.85 &-0.87 &\cellcolor[HTML]{e7f0f1}-0.36 &0.78 &\cellcolor[HTML]{f0f5f6}0.519 \\
25 &0 &\cellcolor[HTML]{dee9eb}0.24 &\cellcolor[HTML]{d8e5e8}-0.12 &\cellcolor[HTML]{e1ebed}-0.26 &1.00 &\cellcolor[HTML]{dae7e9}-0.16 &\cellcolor[HTML]{f3f7f8}-0.54 &-0.85 &-0.91 &\cellcolor[HTML]{eaf1f3}-0.41 &\cellcolor[HTML]{e4edef}0.33 &\cellcolor[HTML]{eef3f4}0.482 \\
27 &0 &\cellcolor[HTML]{d0e0e3}0.00 &\cellcolor[HTML]{d9e6e8}-0.13 &\cellcolor[HTML]{e3ecee}-0.29 &1.00 &\cellcolor[HTML]{e1ebed}-0.26 &\cellcolor[HTML]{f3f7f8}-0.55 &-0.85 &-0.92 &\cellcolor[HTML]{eaf1f3}-0.41 &\cellcolor[HTML]{d0e0e3}0.00 &\cellcolor[HTML]{ebf2f3}0.441 \\
30 &0 &\cellcolor[HTML]{d8e5e8}-0.12 &\cellcolor[HTML]{d9e6e9}-0.14 &\cellcolor[HTML]{e7eff1}-0.36 &0.88 &\cellcolor[HTML]{ebf2f3}-0.42 &\cellcolor[HTML]{f3f7f8}-0.55 &-0.85 &-0.93 &\cellcolor[HTML]{eaf1f3}-0.41 &\cellcolor[HTML]{deeaec}-0.22 &\cellcolor[HTML]{eef4f5}0.488 \\
0 &0.5 &7.24 &3.95 &4.59 &25.75 &58.42 &2.74 &\cellcolor[HTML]{e6eff1}-0.35 &2.80 &2.14 &15.2 &12.3 \\
10 &0.5 &\cellcolor[HTML]{d7e4e7}0.12 &\cellcolor[HTML]{d2e2e4}-0.03 &\cellcolor[HTML]{dfeaec}-0.24 &2.13 &\cellcolor[HTML]{f7fafa}0.63 &\cellcolor[HTML]{eaf1f2}-0.40 &-0.85 &-0.83 &\cellcolor[HTML]{d9e6e9}-0.14 &0.78 &\cellcolor[HTML]{f6f9f9}0.613 \\
15 &0.5 &\cellcolor[HTML]{d8e5e8}-0.12 &\cellcolor[HTML]{d5e3e6}-0.06 &\cellcolor[HTML]{e3edef}-0.30 &1.25 &\cellcolor[HTML]{d9e6e8}0.16 &\cellcolor[HTML]{eff5f6}-0.49 &-0.87 &-0.87 &\cellcolor[HTML]{dfeaec}-0.23 &\cellcolor[HTML]{dde9eb}0.22 &\cellcolor[HTML]{ecf2f4}0.457 \\
18 &0.5 &\cellcolor[HTML]{dce8ea}-0.18 &\cellcolor[HTML]{d6e4e7}-0.08 &\cellcolor[HTML]{e5eef0}-0.32 &1.13 &\cellcolor[HTML]{d4e3e5}-0.05 &\cellcolor[HTML]{f0f5f6}-0.50 &-0.87 &-0.88 &\cellcolor[HTML]{e2ecee}-0.27 &\cellcolor[HTML]{d6e4e7}0.11 &\cellcolor[HTML]{ebf2f3}0.440 \\
20 &0.5 &\cellcolor[HTML]{dce8ea}-0.18 &\cellcolor[HTML]{d6e4e7}-0.09 &\cellcolor[HTML]{e6eef0}-0.34 &1.13 &\cellcolor[HTML]{d7e5e7}-0.11 &\cellcolor[HTML]{f0f5f6}-0.50 &-0.90 &-0.88 &\cellcolor[HTML]{e2ecee}-0.27 &\cellcolor[HTML]{d0e0e3}0.00 &\cellcolor[HTML]{ebf2f3}0.439 \\
22 &0.5 &\cellcolor[HTML]{dfeaec}-0.24 &\cellcolor[HTML]{d7e5e7}-0.10 &\cellcolor[HTML]{e6eff1}-0.35 &1.13 &\cellcolor[HTML]{d7e5e7}-0.11 &\cellcolor[HTML]{f1f6f7}-0.52 &-0.90 &-0.89 &\cellcolor[HTML]{e4eeef}-0.32 &\cellcolor[HTML]{d7e5e8}-0.11 &\cellcolor[HTML]{edf3f4}0.466 \\
25 &0.5 &\cellcolor[HTML]{e3edee}-0.29 &\cellcolor[HTML]{d7e5e8}-0.11 &\cellcolor[HTML]{e8f0f1}-0.37 &1.13 &\cellcolor[HTML]{e1ebed}-0.26 &\cellcolor[HTML]{f3f7f8}-0.54 &-0.90 &-0.90 &\cellcolor[HTML]{e7f0f1}-0.36 &\cellcolor[HTML]{deeaec}-0.22 &\cellcolor[HTML]{eff5f6}0.510 \\
27 &0.5 &\cellcolor[HTML]{eaf2f3}-0.41 &\cellcolor[HTML]{d8e6e8}-0.12 &\cellcolor[HTML]{e8f0f1}-0.37 &1.13 &\cellcolor[HTML]{e4eeef}-0.32 &\cellcolor[HTML]{f3f7f8}-0.55 &-0.90 &-0.91 &\cellcolor[HTML]{e7f0f1}-0.36 &\cellcolor[HTML]{e5eef0}-0.33 &\cellcolor[HTML]{f1f6f7}0.540 \\
30 &0.5 &\cellcolor[HTML]{eef4f5}-0.47 &\cellcolor[HTML]{d9e6e8}-0.13 &\cellcolor[HTML]{ebf2f3}-0.42 &0.88 &\cellcolor[HTML]{eef4f5}-0.47 &\cellcolor[HTML]{f3f7f8}-0.55 &-0.90 &-0.91 &\cellcolor[HTML]{e7f0f1}-0.36 &\cellcolor[HTML]{e5eef0}-0.33 &\cellcolor[HTML]{f2f6f7}0.544 \\
0 &0.9 &\cellcolor[HTML]{dbe7e9}0.18 &1.23 &\cellcolor[HTML]{dee9eb}0.23 &8.00 &9.26 &\cellcolor[HTML]{eaf1f2}0.42 &-0.90 &\cellcolor[HTML]{e7f0f1}-0.37 &\cellcolor[HTML]{ecf2f3}0.45 &1.67 &2.27 \\
10 &0.9 &\cellcolor[HTML]{eaf2f3}-0.41 &\cellcolor[HTML]{d3e2e5}-0.05 &\cellcolor[HTML]{e5eef0}-0.32 &1.63 &\cellcolor[HTML]{dde8ea}0.21 &\cellcolor[HTML]{eff5f6}-0.49 &-0.90 &-0.87 &\cellcolor[HTML]{e4eeef}-0.32 &\cellcolor[HTML]{ecf3f4}-0.44 &\cellcolor[HTML]{f3f7f8}0.564 \\
15 &0.9 &\cellcolor[HTML]{eef4f5}-0.47 &\cellcolor[HTML]{d5e4e6}-0.08 &\cellcolor[HTML]{e6eff1}-0.35 &1.38 &\cellcolor[HTML]{d7e5e7}-0.11 &\cellcolor[HTML]{f1f6f7}-0.52 &-0.90 &-0.91 &\cellcolor[HTML]{e7f0f1}-0.36 &\cellcolor[HTML]{f3f7f8}-0.56 &\cellcolor[HTML]{f3f7f8}0.563 \\
18 &0.9 &\cellcolor[HTML]{f2f6f7}-0.53 &\cellcolor[HTML]{d7e5e7}-0.10 &\cellcolor[HTML]{e9f1f2}-0.39 &1.25 &\cellcolor[HTML]{e1ebed}-0.26 &\cellcolor[HTML]{f2f7f7}-0.53 &-0.90 &-0.91 &\cellcolor[HTML]{e7f0f1}-0.36 &\cellcolor[HTML]{f3f7f8}-0.56 &\cellcolor[HTML]{f4f7f8}0.579 \\
20 &0.9 &\cellcolor[HTML]{f5f9f9}-0.59 &\cellcolor[HTML]{d7e5e7}-0.10 &\cellcolor[HTML]{e9f1f2}-0.39 &1.25 &\cellcolor[HTML]{e8f0f1}-0.37 &\cellcolor[HTML]{f3f7f8}-0.54 &-0.90 &-0.91 &\cellcolor[HTML]{e7f0f1}-0.36 &\cellcolor[HTML]{f3f7f8}-0.56 &\cellcolor[HTML]{f5f8f9}0.597 \\
22 &0.9 &\cellcolor[HTML]{f5f9f9}-0.59 &\cellcolor[HTML]{d7e5e7}-0.10 &\cellcolor[HTML]{ebf2f3}-0.42 &1.13 &\cellcolor[HTML]{ebf2f3}-0.42 &\cellcolor[HTML]{f3f7f8}-0.55 &-0.92 &-0.91 &\cellcolor[HTML]{e7f0f1}-0.36 &\cellcolor[HTML]{f3f7f8}-0.56 &\cellcolor[HTML]{f5f8f9}0.596 \\
25 &0.9 &\cellcolor[HTML]{f5f9f9}-0.59 &\cellcolor[HTML]{d7e5e8}-0.11 &\cellcolor[HTML]{edf3f4}-0.45 &1.13 &\cellcolor[HTML]{f5f8f9}-0.58 &\cellcolor[HTML]{f4f8f8}-0.56 &-0.92 &-0.92 &\cellcolor[HTML]{eaf1f3}-0.41 &\cellcolor[HTML]{f3f7f8}-0.56 &\cellcolor[HTML]{f7f9fa}0.623 \\
27 &0.9 &\cellcolor[HTML]{f5f9f9}-0.59 &\cellcolor[HTML]{d8e6e8}-0.12 &\cellcolor[HTML]{eef4f5}-0.47 &1.13 &\cellcolor[HTML]{f5f8f9}-0.58 &\cellcolor[HTML]{f4f8f8}-0.56 &-0.92 &-0.93 &\cellcolor[HTML]{eaf1f3}-0.41 &\cellcolor[HTML]{f3f7f8}-0.56 &\cellcolor[HTML]{f7f9fa}0.627 \\
30 &0.9 &\cellcolor[HTML]{f9fbfc}-0.65 &\cellcolor[HTML]{d9e6e8}-0.13 &\cellcolor[HTML]{eff5f6}-0.49 &0.88 &\cellcolor[HTML]{fbfdfd}-0.68 &\cellcolor[HTML]{f5f9f9}-0.58 &-0.92 &-0.93 &\cellcolor[HTML]{edf3f4}-0.45 &\cellcolor[HTML]{f3f7f8}-0.56 &\cellcolor[HTML]{f7f9fa}0.627 \\
\bottomrule
\end{tabular}
}
\caption{Relative errors of our counting method with different settings across the validation set's videos. $\gamma$ represents the threshold for number of frames per track ID to count track. $\tau$ represents detection confidence score threshold. Darker color indicates better performance. Note that we include the sign for per species errors to indicate over (postive) or under (negative) counting, but the absolute values of relative error are used in the mean computation.}\label{tab:val-err}

\end{table*}

\begin{table*}[ht!]\centering
\scriptsize
\resizebox{\linewidth}{!}{
\begin{tabular}{lrrrrrrrrrrrrrr}\toprule
\multicolumn{12}{c}{test set per species errors} &test &val \\\cmidrule{1-14}
\textbf{$\gamma$} &\textbf{$\tau$} &BS &FPU &GG &LLS &RSG &SL &LS &WSSC &WSpSC &YG &mean &mean \\
0 &0 &6.00 &4.73 &15.38 &46.66 &70.23 &3.29 &2.57 &2.84 &3.71 &12.21 &16.8 &13.5 \\
10 &0 &\cellcolor[HTML]{dce7ea}0.19 &\cellcolor[HTML]{e4edef}0.33 &1.22 &3.62 &2.02 &\cellcolor[HTML]{e4eeef}-0.32 &\cellcolor[HTML]{edf3f4}-0.45 &-0.77 &\cellcolor[HTML]{d0e0e3}0.00 &1.08 &1.00 &1.09 \\
15 &0 &\cellcolor[HTML]{dae7e9}-0.15 &\cellcolor[HTML]{dce8ea}0.20 &\cellcolor[HTML]{ebf2f3}0.44 &2.03 &\cellcolor[HTML]{dae6e9}0.17 &\cellcolor[HTML]{ecf2f4}-0.44 &\cellcolor[HTML]{f9fbfb}-0.64 &-0.85 &\cellcolor[HTML]{d4e3e6}-0.06 &\cellcolor[HTML]{e7eff0}0.37 &\cellcolor[HTML]{f1f6f6}0.535 &\cellcolor[HTML]{f8fafa}0.641 \\
18 &0 &\cellcolor[HTML]{e0ebed}-0.25 &\cellcolor[HTML]{dae7e9}0.17 &\cellcolor[HTML]{e3ecee}0.31 &1.52 &\cellcolor[HTML]{dee9eb}-0.21 &\cellcolor[HTML]{eff4f6}-0.48 &\cellcolor[HTML]{fdfefe}-0.71 &-0.89 &\cellcolor[HTML]{dce8ea}-0.18 &\cellcolor[HTML]{d1e1e3}0.03 &\cellcolor[HTML]{edf3f4}0.473 &\cellcolor[HTML]{f3f7f7}0.560 \\
20 &0 &\cellcolor[HTML]{e6eff0}-0.35 &\cellcolor[HTML]{d9e6e8}0.16 &\cellcolor[HTML]{dee9eb}0.23 &1.34 &\cellcolor[HTML]{e7eff1}-0.35 &\cellcolor[HTML]{f0f6f6}-0.51 &-0.77 &-0.91 &\cellcolor[HTML]{dfeaec}-0.24 &\cellcolor[HTML]{d9e6e8}-0.13 &\cellcolor[HTML]{eff4f5}0.499 &\cellcolor[HTML]{f1f5f6}0.531 \\
22 &0 &\cellcolor[HTML]{e9f0f2}-0.38 &\cellcolor[HTML]{d8e5e8}0.14 &\cellcolor[HTML]{d8e5e7}0.13 &1.07 &\cellcolor[HTML]{ecf3f4}-0.44 &\cellcolor[HTML]{f2f6f7}-0.53 &-0.80 &-0.91 &\cellcolor[HTML]{dfeaec}-0.24 &\cellcolor[HTML]{dce8ea}-0.18 &\cellcolor[HTML]{eef3f4}0.482 &\cellcolor[HTML]{f0f5f6}0.519 \\
25 &0 &\cellcolor[HTML]{ecf3f4}-0.44 &\cellcolor[HTML]{d6e4e6}0.10 &\cellcolor[HTML]{d1e1e4}-0.01 &0.93 &\cellcolor[HTML]{f4f8f8}-0.56 &\cellcolor[HTML]{f3f7f8}-0.54 &-0.80 &-0.92 &\cellcolor[HTML]{dfeaec}-0.24 &\cellcolor[HTML]{e1ebed}-0.26 &\cellcolor[HTML]{eef3f4}0.481 &\cellcolor[HTML]{eef3f4}0.482 \\
27 &0 &\cellcolor[HTML]{edf4f5}-0.46 &\cellcolor[HTML]{d5e3e6}0.09 &\cellcolor[HTML]{d5e3e6}-0.06 &0.79 &\cellcolor[HTML]{f4f8f8}-0.56 &\cellcolor[HTML]{f3f7f8}-0.55 &-0.83 &-0.93 &\cellcolor[HTML]{dfeaec}-0.24 &\cellcolor[HTML]{e3ecee}-0.29 &\cellcolor[HTML]{eef3f4}0.480 &\cellcolor[HTML]{ebf2f3}0.441 \\
30 &0 &\cellcolor[HTML]{f6f9fa}-0.60 &\cellcolor[HTML]{d4e2e5}0.07 &\cellcolor[HTML]{d7e5e7}-0.10 &\cellcolor[HTML]{f6f9fa}0.62 &\cellcolor[HTML]{fafcfc}-0.67 &\cellcolor[HTML]{f4f8f9}-0.57 &-0.84 &-0.93 &\cellcolor[HTML]{e7eff1}-0.35 &\cellcolor[HTML]{e6eff0}-0.34 &\cellcolor[HTML]{eff5f6}0.509 &\cellcolor[HTML]{eef4f5}0.488 \\
0 &0.5 &4.90 &4.24 &11.88 &44.66 &66.31 &2.82 &1.15 &2.59 &3.82 &9.26 &15.2 &12.3 \\
10 &0.5 &\cellcolor[HTML]{ecf3f4}-0.44 &\cellcolor[HTML]{dfe9eb}0.24 &\cellcolor[HTML]{e6eef0}0.36 &2.79 &\cellcolor[HTML]{f8fafb}0.65 &\cellcolor[HTML]{ebf2f3}-0.42 &\cellcolor[HTML]{fdfefe}-0.71 &-0.82 &\cellcolor[HTML]{dce8ea}-0.18 &\cellcolor[HTML]{d1e1e3}0.03 &\cellcolor[HTML]{f9fbfb}0.663 &\cellcolor[HTML]{f6f9f9}0.613 \\
15 &0.5 &\cellcolor[HTML]{f1f6f7}-0.52 &\cellcolor[HTML]{dae7e9}0.17 &\cellcolor[HTML]{d4e2e5}0.06 &1.79 &\cellcolor[HTML]{d0e0e3}0.00 &\cellcolor[HTML]{eef4f5}-0.47 &-0.80 &-0.85 &\cellcolor[HTML]{dce8ea}-0.18 &\cellcolor[HTML]{e3ecee}-0.29 &\cellcolor[HTML]{f0f5f6}0.514 &\cellcolor[HTML]{ecf2f4}0.457 \\
18 &0.5 &\cellcolor[HTML]{f2f7f8}-0.54 &\cellcolor[HTML]{d9e6e8}0.15 &\cellcolor[HTML]{d1e1e4}-0.01 &1.34 &\cellcolor[HTML]{d8e6e8}-0.12 &\cellcolor[HTML]{f0f5f6}-0.50 &-0.84 &-0.89 &\cellcolor[HTML]{dfeaec}-0.24 &\cellcolor[HTML]{e8f0f1}-0.37 &\cellcolor[HTML]{eff4f5}\textbf{0.500} &\cellcolor[HTML]{ebf2f3}0.440 \\
20 &0.5 &\cellcolor[HTML]{f3f8f8}-0.56 &\cellcolor[HTML]{d8e5e8}0.14 &\cellcolor[HTML]{d2e2e4}-0.03 &1.28 &\cellcolor[HTML]{e0ebed}-0.25 &\cellcolor[HTML]{f1f6f7}-0.51 &-0.84 &-0.91 &\cellcolor[HTML]{dfeaec}-0.24 &\cellcolor[HTML]{e9f1f2}-0.39 &\cellcolor[HTML]{f0f5f6}0.515 &\cellcolor[HTML]{ebf2f3}0.439 \\
22 &0.5 &\cellcolor[HTML]{f3f8f8}-0.56 &\cellcolor[HTML]{d8e5e7}0.13 &\cellcolor[HTML]{d5e4e6}-0.08 &1.03 &\cellcolor[HTML]{e3edee}-0.29 &\cellcolor[HTML]{f2f7f7}-0.53 &-0.84 &-0.91 &\cellcolor[HTML]{dfeaec}-0.24 &\cellcolor[HTML]{e9f1f2}-0.39 &\cellcolor[HTML]{eff4f5}0.501 &\cellcolor[HTML]{edf3f4}0.466 \\
25 &0.5 &\cellcolor[HTML]{f5f8f9}-0.58 &\cellcolor[HTML]{d6e4e7}0.11 &\cellcolor[HTML]{d9e6e8}-0.13 &0.93 &\cellcolor[HTML]{ebf2f3}-0.42 &\cellcolor[HTML]{f2f7f8}-0.54 &-0.84 &-0.92 &\cellcolor[HTML]{dfeaec}-0.24 &\cellcolor[HTML]{ebf2f3}-0.42 &\cellcolor[HTML]{f0f5f6}0.512 &\cellcolor[HTML]{eff5f6}0.510 \\
27 &0.5 &\cellcolor[HTML]{f6f9fa}-0.60 &\cellcolor[HTML]{d6e4e6}0.10 &\cellcolor[HTML]{dae7e9}-0.15 &0.86 &\cellcolor[HTML]{edf3f5}-0.46 &\cellcolor[HTML]{f3f8f8}-0.56 &-0.84 &-0.93 &\cellcolor[HTML]{dfeaec}-0.24 &\cellcolor[HTML]{edf3f4}-0.45 &\cellcolor[HTML]{f0f5f6}0.518 &\cellcolor[HTML]{f1f6f7}0.540 \\
30 &0.5 &\cellcolor[HTML]{f7fafa}-0.62 &\cellcolor[HTML]{d4e3e5}0.08 &\cellcolor[HTML]{dce8ea}-0.18 &\cellcolor[HTML]{f9fbfb}0.66 &\cellcolor[HTML]{f1f6f7}-0.52 &\cellcolor[HTML]{f4f8f9}-0.57 &-0.85 &-0.94 &\cellcolor[HTML]{e7eff1}-0.35 &\cellcolor[HTML]{edf3f4}-0.45 &\cellcolor[HTML]{f0f5f6}0.521 &\cellcolor[HTML]{f2f6f7}0.544 \\
0 &0.9 &\cellcolor[HTML]{ecf3f4}-0.44 &1.51 &0.79 &13.34 &11.00 &\cellcolor[HTML]{edf3f4}0.48 &-0.79 &\cellcolor[HTML]{e7eff1}-0.36 &0.82 &\cellcolor[HTML]{fcfdfd}0.71 &3.03 &2.27 \\
10 &0.9 &\cellcolor[HTML]{fdfefe}-0.71 &\cellcolor[HTML]{dae6e9}0.17 &\cellcolor[HTML]{dbe7ea}-0.17 &1.90 &\cellcolor[HTML]{d2e1e4}-0.02 &\cellcolor[HTML]{eff5f6}-0.49 &-0.88 &-0.89 &\cellcolor[HTML]{eaf2f3}-0.41 &\cellcolor[HTML]{f1f6f7}-0.53 &\cellcolor[HTML]{f6f9f9}0.616 &\cellcolor[HTML]{f3f7f8}0.564 \\
15 &0.9 &\cellcolor[HTML]{feffff}-0.73 &\cellcolor[HTML]{d7e4e7}0.12 &\cellcolor[HTML]{e0ebed}-0.24 &1.07 &\cellcolor[HTML]{ebf2f3}-0.42 &\cellcolor[HTML]{f3f7f8}-0.55 &-0.88 &-0.91 &\cellcolor[HTML]{eaf2f3}-0.41 &\cellcolor[HTML]{f8fbfb}-0.63 &\cellcolor[HTML]{f5f8f9}0.596 &\cellcolor[HTML]{f3f7f8}0.563 \\
18 &0.9 &-0.75 &\cellcolor[HTML]{d6e4e6}0.10 &\cellcolor[HTML]{e2ecee}-0.28 &0.76 &\cellcolor[HTML]{f2f7f8}-0.54 &\cellcolor[HTML]{f4f8f9}-0.57 &-0.88 &-0.93 &\cellcolor[HTML]{eef4f5}-0.47 &\cellcolor[HTML]{fdfefe}-0.71 &\cellcolor[HTML]{f5f8f9}0.599 &\cellcolor[HTML]{f4f7f8}0.579 \\
20 &0.9 &-0.75 &\cellcolor[HTML]{d5e3e6}0.09 &\cellcolor[HTML]{e4edef}-0.31 &\cellcolor[HTML]{f9fbfb}0.66 &\cellcolor[HTML]{f4f8f8}-0.56 &\cellcolor[HTML]{f5f9f9}-0.58 &-0.88 &-0.94 &\cellcolor[HTML]{eef4f5}-0.47 &\cellcolor[HTML]{fdfefe}-0.71 &\cellcolor[HTML]{f5f8f9}0.595 &\cellcolor[HTML]{f5f8f9}0.597 \\
22 &0.9 &-0.75 &\cellcolor[HTML]{d5e3e6}0.08 &\cellcolor[HTML]{e4edef}-0.31 &\cellcolor[HTML]{f2f6f7}0.55 &\cellcolor[HTML]{f5f9f9}-0.58 &\cellcolor[HTML]{f5f9fa}-0.59 &-0.89 &-0.95 &\cellcolor[HTML]{eef4f5}-0.47 &-0.74 &\cellcolor[HTML]{f5f8f9}0.591 &\cellcolor[HTML]{f5f8f9}0.596 \\
25 &0.9 &-0.75 &\cellcolor[HTML]{d3e2e5}0.06 &\cellcolor[HTML]{e6eff0}-0.35 &\cellcolor[HTML]{e9f1f2}0.41 &\cellcolor[HTML]{fafcfc}-0.67 &\cellcolor[HTML]{f6f9fa}-0.60 &-0.92 &-0.95 &\cellcolor[HTML]{eef4f5}-0.47 &-0.76 &\cellcolor[HTML]{f5f8f9}0.594 &\cellcolor[HTML]{f7f9fa}0.623 \\
27 &0.9 &-0.75 &\cellcolor[HTML]{d2e1e4}0.05 &\cellcolor[HTML]{e6eff0}-0.35 &\cellcolor[HTML]{e9f1f2}0.41 &\cellcolor[HTML]{fdfefe}-0.71 &\cellcolor[HTML]{f6f9fa}-0.60 &-0.93 &-0.96 &\cellcolor[HTML]{eef4f5}-0.47 &-0.79 &\cellcolor[HTML]{f5f8f9}0.602 &\cellcolor[HTML]{f7f9fa}0.627 \\
30 &0.9 &-0.75 &\cellcolor[HTML]{d1e1e4}0.03 &\cellcolor[HTML]{e7eff1}-0.36 &\cellcolor[HTML]{dce8ea}0.21 &-0.75 &\cellcolor[HTML]{f7fafa}-0.61 &-0.93 &-0.96 &\cellcolor[HTML]{f2f6f7}-0.53 &-0.82 &\cellcolor[HTML]{f5f8f9}0.595 &\cellcolor[HTML]{f7f9fa}0.627 \\
\bottomrule
\end{tabular}
}

\caption{Relative errors of our counting method with different settings across the test set's videos}\label{tab:test-err}

\end{table*}

\clearpage
\newpage

\end{appendices}



\end{document}